\documentclass{article}

\usepackage{PRIMEarxiv}

\usepackage[utf8]{inputenc} % allow utf-8 input
\usepackage[T1]{fontenc}    % use 8-bit T1 fonts
\usepackage{hyperref}       % hyperlinks
\usepackage{url}            % simple URL typesetting
\usepackage{booktabs}       % professional-quality tables
\usepackage{amsfonts}       % blackboard math symbols
\usepackage{nicefrac}       % compact symbols for 1/2, etc.
\usepackage{microtype}      % microtypography
\usepackage{lipsum}
\usepackage{fancyhdr}       % header
\usepackage{graphicx}       % graphics
\graphicspath{{media/}}     % organize your images and other figures under media/ folder
\usepackage{nicefrac}       % compact symbols for 1/2, etc.
\usepackage{microtype}      % microtypography
\usepackage{xcolor}         % colors
\usepackage{subfigure}
\usepackage{wrapfig}
\usepackage{enumitem}
\usepackage{booktabs}

\usepackage{amsmath}
\usepackage{amssymb}
\usepackage{mathtools}
\usepackage{amsthm}
\usepackage{bbm}
\usepackage{array}

\title{On the Inflation of KNN-Shapley Value
}

\author{
    Ziao~Yang\footnotemark[1]~,~~~~Han~Yue\footnotemark[1]~,~~~~Jian~Chen\footnotemark[2]~,~~~~and~Hongfu~Liu\footnotemark[1] \\
  Brandeis University\footnotemark[1]~,~~~~Tsinghua University\footnotemark[2]\\
  \{ziaoyang,hanyue,hongfuliu\}@brandeis.edu\footnotemark[1]~,~~~~chenj@sem.tsinghua.edu.cn\footnotemark[2]
}

\begin{document}
\maketitle

\begin{abstract}
Shapley value-based data valuation methods, originating from cooperative game theory, quantify the usefulness of each individual sample by considering its contribution to all possible training subsets. Despite their extensive applications, these methods encounter the challenge of value inflation—while samples with negative Shapley values are detrimental, some with positive values can also be harmful. This challenge prompts two fundamental questions: the suitability of zero as a threshold for distinguishing detrimental from beneficial samples and the determination of an appropriate threshold. To address these questions, we focus on KNN-Shapley and propose Calibrated KNN-Shapley (CKNN-Shapley), which calibrates zero as the threshold to distinguish detrimental samples from beneficial ones by mitigating the negative effects of small-sized training subsets. Through extensive experiments, we demonstrate the effectiveness of CKNN-Shapley in alleviating data valuation inflation, detecting detrimental samples, and assessing data quality. We also extend our approach beyond conventional classification settings, applying it to diverse and practical scenarios such as learning with mislabeled data, online learning with stream data, and active learning for label annotation.
\end{abstract}

% keywords can be removed
% \keywords{First keyword \and Second keyword \and More}

\section{Introduction}
The significance of data as inputs in machine learning algorithms cannot be overstated for algorithmic performance~\cite{liang2022advances,zha2023data}. Beyond the development of sophisticated algorithms, there is a growing emphasis on the curation of high-quality training sets. Departing from the conventional practice of isolating data from the learning algorithm, a surge in data-centric learning has emerged to assess the valuation of data within the context of a learning task. This encompasses a spectrum of tasks, ranging from outlier detection~\cite{boukerche2020outlier} to noisy label correction~\cite{zheng2021meta}, best subset selection~\cite{hazimeh2020fast} to sample reweighting~\cite{li2022achieving}, and antidote data generation~\cite{chhabra2022fair, li2022learning} to active labeling~\cite{tharwat2023survey,liu2021influence}.

Understanding the value of an individual data sample is fundamental in data-centric learning, often assessed by the shift in learning utility when the model is trained with and without that specific sample. Leave-one-out influence~\cite{cook1982residuals}, a straightforward method, offers an initial assessment of the relative influence of the specific sample compared to the rest of the training set. On the other hand, Shapley value-based data valuation methods~\cite{shapley1953value,roth1988shapley}, originating from cooperative game theory, quantifies the usefulness of each individual sample towards the utility on a validation set by considering its contribution to all possible training subsets. Unlike the leave-one-out influence, Shapley value represents the weighted average utility change resulting from adding the point to different training subsets, showcasing greater robustness in diverse contexts~\cite{bordt2023shapley,tsai2023faith,sundararajan2020shapley}. Despite the absence of assumptions on the learning model, the retraining-based category mentioned above incurs significant computation costs, especially for large-scale data analysis and deep models~\cite{hammoudeh2022training}.

With the advent of KNN-Shapley~\cite{jia2019towards,jia2021scalability}, a pragmatic tool enabling the computation of Shapley values without the need for costly model retraining, Shapley-based approaches have become feasible and widely applied. KNN-Shapley leverages the K-Nearest Neighbors classifier as a surrogate for the original learning model, recursively calculating the Shapley value for each training sample. Despite their promise, KNN-Shapley and its variants grapple with the issue of value inflation. While samples with negative Shapley values are recognized as detrimental, the dilemma arises as certain samples with positive values may also have harmful effects. This challenge gives rise to two pivotal questions: the appropriateness of zero as a threshold for distinguishing detrimental from beneficial samples and the identification of a suitable threshold.

\textbf{Our Contributions.} In this paper, we focus on addressing the value inflation issue observed in KNN-Shapley. To tackle this, we propose Calibrated KNN-Shapley (CKNN-Shapley), which calibrates zero as the threshold to distinguish detrimental samples from beneficial ones by mitigating the negative effects of small training subsets. Our contributions are summarized as follows:\vspace{-1mm}
\begin{itemize}[wide=10pt, leftmargin=*, nosep]
    \item We unveil the value inflation issue of KNN-Shapley, which not only misidentifies a portion of detrimental samples as beneficial, but also distorts the interpretation of data valuation. Beyond the misidentified samples, value inflation further impacts the assessment of beneficial samples.\vspace{1mm}
    \item We propose CKNN-Shapley to calibrate zero as the threshold to distinguish detrimental samples from beneficial ones. Our hypothesis attributes value inflation to improper subset selection in KNN-Shapley, and we address this issue through a straightforward yet effective strategy by imposing a constraint on the size of training subsets.\vspace{1mm}
    \item We perform comprehensive experiments on various benchmark datasets, comparing CKNN-Shapley with KNN-Shapley-based methods. The results showcase the effectiveness of CKNN-Shapley in mitigating value inflation and improving classification performance. We also extend our approach beyond conventional classification settings, applying it to learning with mislabeled data, online learning with stream data, and active learning for label annotation.
\end{itemize}

\textbf{Related Work}. We introduce Shapley-based methods with a major focus on data valuation and KNN-Shapley. \textit{(i) Shapley-based Data Valuation}. The Shapley value~\cite{shapley1953value,roth1988shapley} measures the weighted average utility change when adding a point to all possible training subsets, making it a primary tool for assessing the valuation of individual samples~\cite{jiang2023opendataval}. Shapley-value based methods have found extensive application in various domains, including variable selection~\cite{cohen2005feature,zaeri2018feature}, feature importance~\cite{lundberg2017unified,covert2020improving,jethani2021fastshap}, model valuation~\cite{rozemberczki2021shapley}, health care~\cite{pandl2021trustworthy,tang2021data}, federated learning~\cite{han2021data}, collaborative learning~\cite{sim2020collaborative}, data debugging~\cite{deutch2021explanations}, and distribution analysis~\cite{schoch2022cs,ghorbani2020distributional}.
Building upon this concept, Beta Shapley~\cite{kwon2022beta} and Banzhaf value~\cite{wang2023data} are developed by relaxing the efficiency axiom of the Shapley value. While Shapley-based data valuation approaches are model-agnostic, exponential model retraining renders these methods computationally challenging even for small datasets ~\cite{ghorbani2019data,jia2019towards}. Efforts to accelerate computation include efficient sampling~\cite{zhang2023efficient}, utility learning~\cite{wang2021improving}, and the assumption of independent utility~\cite{luo2022shapley}. In addition to data valuation. \textit{(ii) KNN-Shapley Data Valuation}. KNN-Shapley~\cite{jia2018efficient,wang2023note} emerges as one of the most promising solutions to mitigate the computational challenges associated with Shapley values. It leverages assumptions about the learning model by employing a KNN classifier as a surrogate, recursively computing the Shapley value for each training sample without the need for retraining. This high efficiency has spurred the development of numerous KNN-Shapley variants, such as weighted KNN~\cite{wang2024efficient}, soft KNN~\cite{wang2023note}, and threshold KNN\cite{wang2023threshold}, tailored to enhance generalization, sample reuse, and privacy risks, respectively. In this paper, our focus is on the KNN-Shapley category, with the objective of addressing issues related to value inflation.

\section{Preliminaries and Motivation}
In this section, we introduce the preliminaries of Shapley-based valuation and reveal its value inflation. 

\textbf{Preliminaries}. For a training set $\mathcal{D}$ with $N$ samples and a learning algorithm $\mathcal{A}$, let $U_{\mathcal{A},\mathcal{D}_v}(\mathcal{D})$ represent the model utility with all training data on the validation set $\mathcal{D}_v$. For simplicity, we use $U(\mathcal{D})$ in the following manuscript. The Shapley value~\cite{shapley1953value} of a training sample $z_i$$\in$$D, $$1$$\leq$$i$$\leq$$N$ is defined as follows:
\begin{equation}\label{eq:shapley}
    \nu^s(z_i) = \frac{1}{N}\sum_{\mathcal{S}\subseteq \mathcal{D}\backslash z_i}\frac{1}{{\binom{N-1}{|\mathcal{S}|}}}[U(\mathcal{S}\cup z_i)-U(\mathcal{S})],
\end{equation}
where $\mathcal{S}$ is a training subset. The Shapley value gauges the average contribution of $z_i$ on all possible subsets of $\mathcal{D}$ without $z_i$ from the cooperative game perspective. Note that obtaining the exact Shapley value necessitates $2^N$ model trainings. Despite no assumption of the learning algorithm $\mathcal{A}$, for large models, this process consumes significant computational resources and time. The impracticality of the exact Shapley value's expensive complexity, even with Monte Carlo approximation, becomes evident in large-scale data analysis with substantial models.

KNN-Shapley~\cite{jia2019towards} occurs as a pragmatic tool for computing Shapley values efficiently. It employs a KNN classifier as a surrogate for the learning algorithm $\mathcal{A}$. For a single validation sample $\mathcal{D}_v$$=$$\{z_v\}$, where $z_v$$=$$(x_v,y_v)$ contains $x_v$ in the feature space and $y_v$ in the label space, a KNN classifier sorts the whole training data and identifies $K$ nearest neighbors in the feature space $(z_{\alpha_1}, z_{\alpha_2}, \cdots, z_{\alpha_K})$ to $z_v$, where $\alpha_k$ represents the index of the training samples with the $k$-th neighbor to the validation sample $z_v$. If the predictive confidence is used as the model utility, \textit{i.e.}, $U(\mathcal{S})=\frac{1}{K}\sum_{k=1}^{\textup{min}\{K,|\mathcal{S}|\}}\mathbbm{1}[y_{\alpha_k}$$=$$y_v]$, KNN-Shapley values can be calculated recursively as follows:
\begin{equation}\label{eq:knn}
\small
    \nu^k_{\textup{}}(z_{\alpha_N}) = \frac{\mathbbm{1}[y_{\alpha_N}=y_v]}{N},\ \
    \nu^k_{\textup{}}(z_{\alpha_i}) = \nu^k_{\textup{}}(z_{\alpha_{i+1}})+\frac{\mathbbm{1}[y_{\alpha_i}=y_v]-\mathbbm{1}[y_{\alpha_{i+1}}=y_v]}{\max\{K,i\}}.
\end{equation}
The above Eq.~(\ref{eq:knn}) can be extended to multiple validation samples by summing up the KNN-Shapley value for each validation sample. This results in a time complexity of $\mathcal{O}(N\log N)$ for KNN-Shapley, significantly faster than the vanilla Shapley. It is worth noting that KNN is a lazy classifier without any training process; therefore, within the retrain-based Shapley framework, KNN-Shapley can directly conduct the inference process, significantly reducing the time complexity compared to other training-required classifiers. Moreover, for certain deep models, KNN-Shapley is compatible with the embedding of training/validation samples for data valuation~\cite{jia2019towards}.

\begin{wrapfigure}{r}{0.5\textwidth} \vspace{-4mm}
  \centering
	\includegraphics[width=0.48\textwidth]{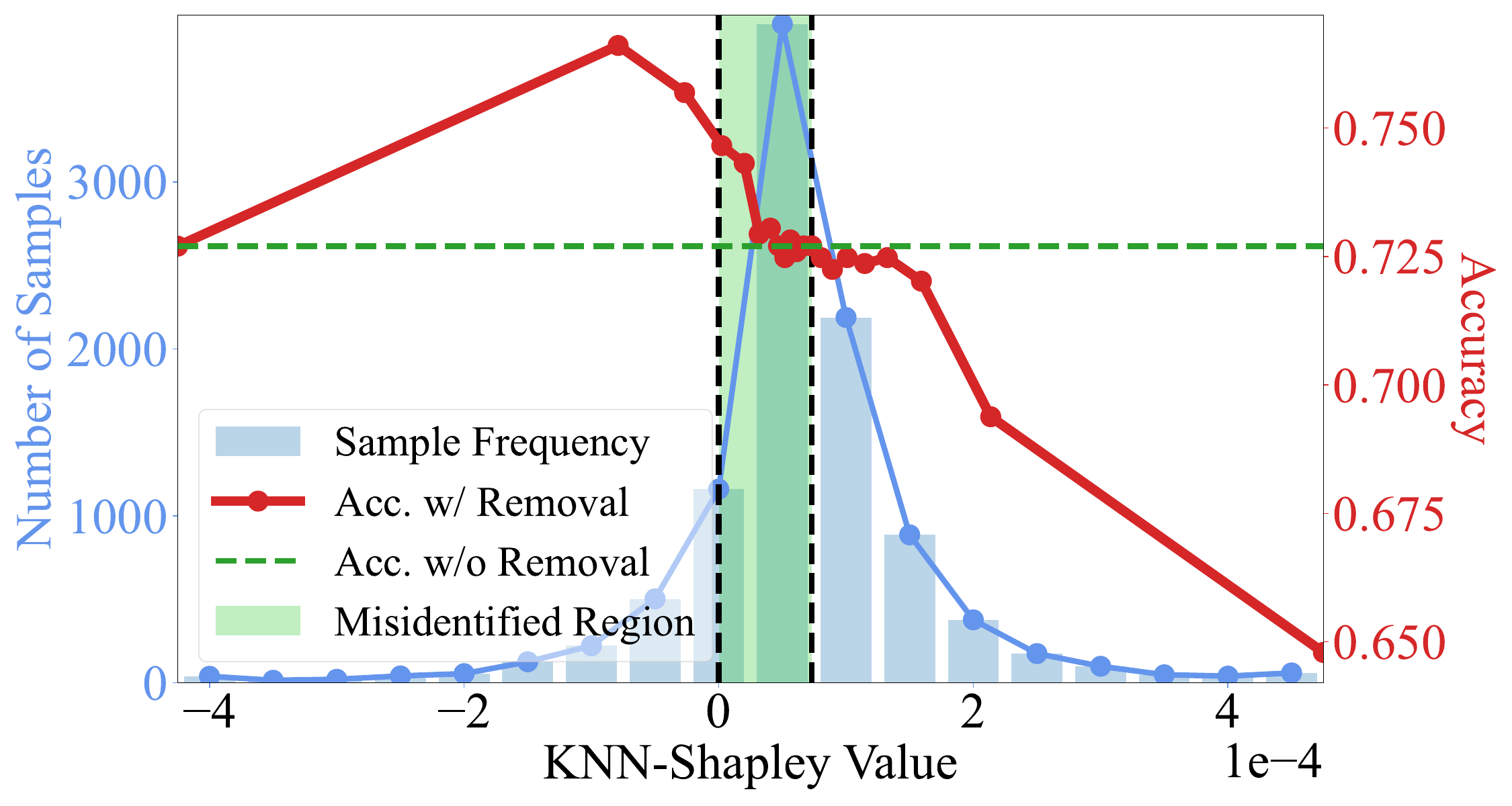}
    \vspace{-2.5mm}
	\caption{Illustration of KNN-Shapley value inflation. The bar plot displays the histogram of KNN-Shapley values for training samples in the \textit{SST-2} dataset~\cite{socher2013recursive}. For the purpose of visualization, we merge the samples with extremely small or large values into the leftmost or rightmost bars. With a segmentation of 20 equally-sized bins based on the ascending order of their values, the red line illustrates KNN performance with a specific bin removed from the training set, while the dashed green line represents KNN performance on the entire training set. By comparing the red and green lines, the detrimental bin can be identified, as the performance improves upon its removal. While samples with negative KNN-Shapley values are generally detrimental, a notable observation is the green shallow highlighted region, where samples are harmful to the learning task despite having positive KNN-Shapley values, indicating the issue of KNN-Shapley value inflation.}\label{fig:motivation}
    \vspace{-2mm}
\end{wrapfigure}
\textbf{Inflation of KNN-Shapley Value}. 
Despite the wide range of applications and efficiency of KNN-Shapley, we have observed a phenomenon of value inflation during its practical usage. This occurs when, contrary to the expectation that samples with negative Shapley values are detrimental, some samples with positive values can also be harmful. Figure~\ref{fig:motivation} illustrates this phenomenon on the \textit{SST-2} dataset~\cite{socher2013recursive}. The bar plot shows the histogram of KNN-Shapley values for training samples. According to the ascending order of their values, we segment the whole training set into 20 equally sized bins. To identify whether samples in a specific bin are detrimental or beneficial, we train the KNN classifier with that bin removed from the training set (denoted by the red line) and compare its performance with the complete training set (denoted by the green line). Due to the equal-size samples in each bin, the markers on the red line do not have uniform intervals and do not align with unevenly sized histograms. While samples with negative KNN-Shapley values are generally detrimental, the highlighted green shallow region reveals a noteworthy observation—samples in this region are harmful to the learning task despite having positive KNN-Shapley values, indicating the issue of KNN-Shapley value inflation. Note that the number of samples with negative KNN-Shapley values is 1,452, while 4,548 samples are in the misidentified region. Moreover, removing all the samples with negative KNN-Shapley values enhances performance from 0.7270 to 0.8160. Conversely, further improvement is achieved by removing samples in the misidentified region, boosting performance from 0.8160 to 0.8980 and underscoring the significance of addressing inflated samples.

Value inflation prompts two
fundamental questions: the suitability of zero as
a threshold for distinguishing detrimental from
beneficial samples and the determination of an
appropriate threshold. To address these questions, we propose Calibrated KNN-Shapley in the next section, which calibrates zero
as the threshold to distinguish detrimental samples
from beneficial ones.

\section{Calibrated KNN-Shapley}
To address the challenge of valuation inflation in KNN-Shapley, we first delve into its underlying mechanism and scrutinize potential factors contributing to negative effects on data valuation. Based on the analyzed reasons, we propose our Calibrated KNN-Shapley by mitigating the negative efforts from improper training subsets.

While the recursive formulation in Eq.~(\ref{eq:knn}) brings about efficient computation, it also introduces inevitable accumulated errors. Specifically, the valuation of a training sample distant from the validation sample can influence other training samples closer to the validation sample, emphasizing the importance of the first term $\nu^k_{\textup{}}(z_{\alpha_N})$. Upon closer examination of $z_{\alpha_N}$, the farthest sample from the validation sample, we observe its minimal impact on the validation sample. Considering the definition of the Shapley value, which measures the average contribution of a training sample across all possible subsets, we discuss two cases. Case I: when $|\mathcal{S}|$$>$$K$, $z_{\alpha_N}$ does not contribute to the utility of the KNN classifier since it is not among the neighbors of $z_v$. Case II: when $|\mathcal{S}|$$\leq$$ K$, $\nu^k_{\textup{}}(z_{\alpha_N})$ is non-negative, \textit{i.e.}, $\nu^k_{\textup{}}(z_{\alpha_N})$$=$$1/N$ if $z_{\alpha_N}$ and $z_v$ share the same label; otherwise, $\nu^k_{\textup{}}(z_{\alpha_N})$$=$$0$. Notably, it is impractical and meaningless to have a training subset with only a few training samples. Furthermore, the number of occurrences in Case I significantly surpasses that in Case II, indicating that Case II is not representative for a learning model and data valuation.

Building on the above analyses, we hypothesize that value inflation stems from improper subset selection in KNN-Shapley. Certain subsets with only a few samples exhibit significant divergence from the original set, leading to an exaggeration of the contribution of a specific sample on these subsets. This, in turn, gives rise to the phenomenon of value inflation. To address the issue of value inflation, we introduce Calibrated KNN-Shapley (CKNN-Shapley) through the selection of suitable training subsets. The training subset in CKNN-Shapley should serve as an effective proxy for the original and complete training set. In this paper, we present a straightforward yet effective strategy by imposing a constraint on the size of training subsets, specifically $|\mathcal{S}|$$\geq$$T$, where $T$ represents the size of the smallest training subsets used to assess the contribution of each training sample. This allows us to compute the CKNN-Shapley value as follows:
\begin{equation}\label{eq:cknn}
\small
\begin{split}
    & \nu^c_{\textup{}}(z_{\alpha_{N}}) = \nu^c_{\textup{}}(z_{\alpha_{N-1}})=\cdots=\nu^c_{\textup{}}(z_{\alpha_{N-T+1}})=0,\\
    &\nu^c_{\textup{}}(z_{\alpha_{N-T}}) = \frac{\mathbbm{1}[y_{\alpha_{N-T}}=y_v]}{N-T},\ \nu^c_{\textup{}}(z_{\alpha_i}) = \nu^c_{\textup{}}(z_{\alpha_{i+1}})+\frac{\mathbbm{1}[y_{\alpha_i}=y_v]-\mathbbm{1}[y_{\alpha_{i+1}}=y_v]}{\max\{K,i\}}.
\end{split}    
\end{equation}
Compared to KNN-Shapley in Eq.~(\ref{eq:knn}), CKNN-Shapley follows a similar recursive fashion but more efficiently. This efficiency stems from directly assigning zero to $T$ samples that are far away from the validation sample. Additionally, in CKNN-Shapley, the selected training subsets consist of at least $T$ samples, preventing the inclusion of improper subsets that could contribute to the valuation. 

\begin{wrapfigure}{r}{0.5\textwidth} 
  \centering\vspace{-8mm}    \centering\includegraphics[width=0.42\textwidth]{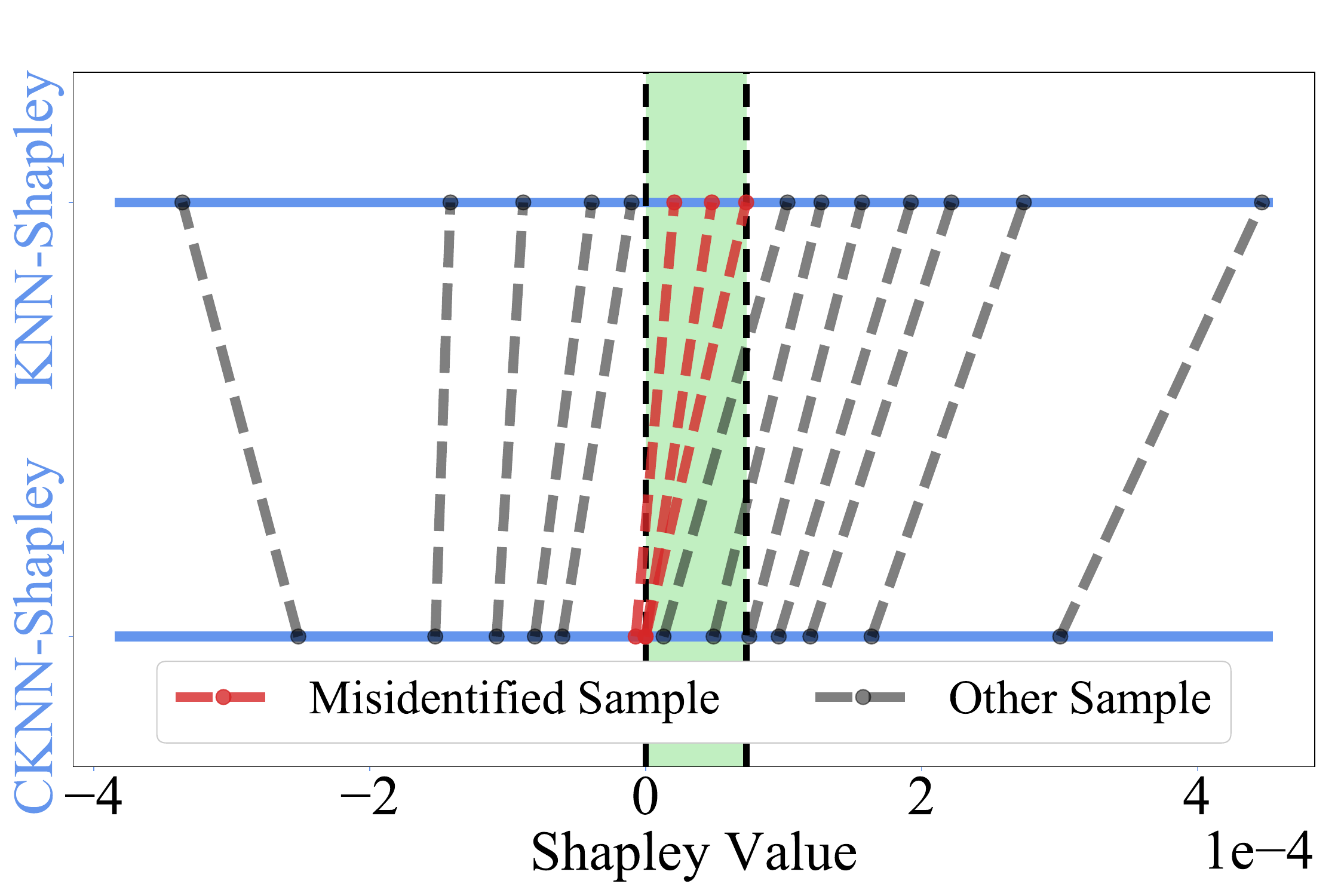}
    \vspace{-3mm}
	\caption{Comparison of KNN-Shapley and CKNN-Shapley value on the \textit{SST-2} dataset~\cite{socher2013recursive}, where each dashed line represents a training sample associated with its KNN-Shapley and CKNN-Shapley values, respectively, and the green region is the misidentified detrimental samples from Figure~\ref{fig:motivation}. The red dashed lines denote the samples that are incorrectly identified by KNN-Shapley but correctly identified by CKNN-Shapley.}\label{fig:comparison}
    \vspace{-5mm}
\end{wrapfigure}
Figure~\ref{fig:comparison} illustrates the comparison between KNN-Shapley and CKNN-Shapley values on the \textit{SST-2} dataset. Each dashed line represents a training sample associated with its KNN-Shapley and CKNN-Shapley values. Detrimental samples misidentified by KNN-Shapley, highlighted by red dashed lines, are assigned negative or zero values in CKNN-Shapley, suggesting that zero in CKNN-Shapley serves as a suitable threshold for distinguishing detrimental from beneficial samples. Additionally, the majority of lines from KNN-Shapley to CKNN-Shapley move from the top right to the bottom left. This indicates that KNN-Shapley not only has a negative effect on the misidentified samples but also tends to inflate the valuation of most samples. This inflation might be attributed to its recursive formulation, leading to the accumulation of errors within KNN-Shapley.

As a variant of KNN-Shapley, our CKNN-Shapley inherently integrates several key axioms from KNN-Shapley, focusing on three fundamental principles: group rationality, fairness, and additivity, as emphasized in~\cite{jia2019towards}. CKNN-Shapley prioritizes fairness, ensuring symmetry (where two identical samples receive identical Shapley values) and zero elements (no contribution, no payment). Additionally, it upholds additivity, where values across multiple utilities sum up to the value under a utility that is the aggregate of all these utilities. Enhancing group rationality, CKNN-Shapley refines this axiom by disregarding subsets with fewer than 
$T$ samples, thereby distributing the utility difference between the entire dataset and the subset. Since the utility of a subset, according to the KNN classifier, expects only the nearest samples to contribute, the utility of such a subset is anticipated to be zero, a factor overlooked in KNN-Shapley. By addressing this, CKNN-Shapley mitigates the adverse impact of overemphasizing the cooperative game with samples distant from the target sample, rendering the axiom of group rationality more compatible with the KNN classifier. 

\begin{table*}[t]\vspace{-1mm}
\centering
\caption{Threshold for distinguishing detrimental from beneficial samples and the misidentification ratio of detrimental samples in Eq.~(\ref{eq:metric}) for KNN-Shapley-based methods. For the first metric, closer to zero is preferable; Regarding the second one, smaller values suggest better performance. The last column takes the absolute values for average. The best results are highlighted in bold. TKNN-Shapley does not return meaningful results on \textit{CIFAR10} with the default parameters, denoted by ``N/A."
}\label{tab:index}\vspace{1mm}
\resizebox{0.98\textwidth}{!}{%
\begin{tabular}{p{3.5cm}|*{9}{>{\centering\arraybackslash}p{1.2cm}}|>{\centering\arraybackslash}p{1.2cm}}
\toprule
Datasets  & \textit{MNIST} & \textit{FMNIST} & \textit{CIFAR10} & \textit{Pol} & \textit{Wind} & \textit{CPU} & \textit{AGnews} & \textit{SST-2} & \textit{News20} & Avg. \\ 
 \midrule
\multicolumn{8}{l}{Threshold for distinguishing detrimental from beneficial samples by $\times$ $1e$$-4$}\\
 \midrule
% LOO$^\#$ & - & \\ 
KNN-Shapley~\cite{jia2018efficient} & 0.2744 & 0.2631 & 0.1632 & \hspace{0.3em}2.0361 & \hspace{0.3em}7.3705 & \hspace{0.3em}6.4458 & 1.6264 & 1.3659 & 1.7215 & 2.3630\\ 
KNN-Shapley-JW~\cite{wang2023note} & 0.2544 & 0.2431 & 0.1432 & \hspace{0.3em}0.5119 & \hspace{0.3em}4.6543 & \hspace{0.3em}3.7429 & 1.3763 & 0.8203 & 1.6780 & 1.4916\\ 
TKNN-Shapley~\cite{wang2023threshold} & 3.8641 & 1.0330 & N/A & -1.4990 & \hspace{0.3em}1.0875 & \hspace{0.3em}5.6960 & \textbf{1.1951} & 0.7143 & 5.2628 & 2.5440\\ 
CKNN-Shapley (Ours) &  \textbf{0.0152} & \textbf{0.0109} & \textbf{0.1163} & \textbf{-0.3059} & \textbf{-0.0245} & \textbf{-2.9412} & 1.1955 & \textbf{0.1604} & \textbf{0.6715} & \textbf{0.6046}\\ 
 \midrule
% LOO$^+$ & - & \\ 
\multicolumn{8}{l}{Misidentification ratio of detrimental samples}\\
 \midrule
KNN-Shapley~\cite{jia2018efficient} & 0.8065 & 0.5145 & 0.4908 & \hspace{0.3em}0.3333 & \hspace{0.3em}0.3864 & \hspace{0.3em}0.5714 & 0.7580 & 0.5600 & 0.6179 & 0.5599\\ 
KNN-Shapley-JW~\cite{wang2023note} & 0.7665 & 0.4367 & 0.3792 & \hspace{0.3em}0.0000 & \hspace{0.3em}0.2727 & \hspace{0.3em}0.3846 & 0.6563 & 0.3421 & 0.5541 & 0.4214\\ 
TKNN-Shapley~\cite{wang2023threshold} & 0.3103 & 1.0000 & N/A & \hspace{0.3em}\textbf{0.0000} & \hspace{0.3em}0.1818 & \hspace{0.3em}0.4118 & 0.2273 & 0.1862 & 0.2000 & 0.3147\\ 
CKNN-Shapley (Ours) & \textbf{0.2000} & \textbf{0.1686} & \textbf{0.0908} & \hspace{0.3em}\textbf{0.0000} & \hspace{0.3em}\textbf{0.0000} & \hspace{0.3em}\textbf{0.0000} & \textbf{0.1538} & \textbf{0.1143} & \textbf{0.1821} & \textbf{0.1011}\\ 
\bottomrule
\end{tabular}
}\vspace{4mm}
\end{table*}

\begin{table*}[t]
\centering
\vspace{-8mm}
\caption{Classification performance of KNN-Shapley-based methods. $\#$ denotes KNN's performance trained on the training set excluding samples with negative valuations, while $+$ represents weighted KNN's performance, where the weights are derived from data valuations.}\label{tab:remove}\vspace{1mm}
\resizebox{0.98\textwidth}{!}{%
\begin{tabular}{p{3.5cm}|*{9}{>{\centering\arraybackslash}p{1.2cm}}|>{\centering\arraybackslash}p{1.2cm}}
\toprule
Method$\backslash$Datasets  & \textit{MNIST} & \textit{FMNIST} & \textit{CIFAR10} & \textit{Pol} & \textit{Wind} & \textit{CPU} & \textit{AGnews} & \textit{SST-2} & \textit{News20} & Avg. \\ 
 \midrule
Vanilla KNN & 0.9630 & 0.8444 & 0.5956 & 0.9400 & 0.8700 & 0.9200 & 0.9060 & 0.7270 & 0.6920 & 0.8287\\
 \midrule
% LOO$^\#$ & - & \\ 
KNN-Shapley~\cite{jia2018efficient}$^\#$ & 0.9682 & 0.8586 & 0.6456 & \textbf{0.9700} & 0.8750 & 0.9650 & 0.9250 & 0.8160 & 0.7580 & 0.8646\\ 
KNN-Shapley-JW~\cite{wang2023note}$^\#$ & 0.9698 & 0.8574 & 0.6514 & \textbf{0.9700} & 0.8650 & 0.9600 & 0.9250 & 0.8498 & 0.7610 & 0.8677\\ 
TKNN-Shapley~\cite{wang2023threshold}$^\#$ & 0.6644 & 0.8356 & N/A  & 0.8150 & 0.8300 & 0.9100 & 0.8990 & 0.7982 & 0.6730 & 0.8032\\ 
CKNN-Shapley (Ours)$^\#$ & \textbf{0.9828} & \textbf{0.8884} & \textbf{0.7164} & \textbf{0.9700} & \textbf{0.9000} & \textbf{0.9700} & \textbf{0.9420} & \textbf{0.8980} & \textbf{0.7960}  & \textbf{0.8960}\\ 
 \midrule
% LOO$^+$ & - & \\ 
KNN-Shapley~\cite{jia2018efficient}$^+$ & 0.9742 & 0.8680 & 0.6598 & 0.9650 & 0.8700 & 0.9300 & 0.9333 & 0.8480 & 0.7790 & 0.8697\\ 
KNN-Shapley-JW~\cite{wang2023note}$^+$ & 0.9754 & 0.8658 & 0.6610 & 0.9650 & 0.8700 & 0.9300 & 0.9320 & 0.8555 & 0.7790 & 0.8704\\ 
TKNN-Shapley~\cite{wang2023threshold}$^+$ & 0.8626 & 0.7900 & N/A & 0.8600 & 0.8200 & 0.8950 & 0.9110 & 0.8326 & 0.7160 & 0.8359\\ 
CKNN-Shapley (Ours)$^+$ & \textbf{0.9890} & \textbf{0.9106} & \textbf{0.7404} & \textbf{0.9700} & \textbf{0.9050} & \textbf{0.9650} & \textbf{0.9450} & \textbf{0.8920} & \textbf{0.8130} & \textbf{0.9033} \\ 
\bottomrule
\end{tabular}%
}\vspace{-4mm}
\end{table*}

\section{Experimental Results}\label{sec:experiment}
In this section, we present the experimental results of KNN-Shapley, its variant TKNN-Shapley, and our proposed CKNN-Shapley on several widely used benchmark datasets in the data valuation domain. We begin by detailing the experimental setup, with a specific emphasis on the evaluation metric used to quantify value inflation. Subsequently, we showcase the algorithmic performance in addressing value inflation and enhancing classification performance. Finally, we delve into several in-depth explorations of CKNN-Shapley for practical applications.

\textbf{Experimental Setup}. Here we use 9 datasets for empirical evaluation. \textit{MNIST}, \textit{FMNIST}~\cite{xiao2017fashion}, and \textit{CIFAR10} are image datasets with 50,000 samples; \textit{Pol}, \textit{Wind}, and \textit{CPU}~\cite{wang2023threshold} are from the telecommunication, meteorology, and computer hardware domains, respectively, with 2,000 samples each; \textit{AGnews}, \textit{SST-2}~\cite{socher2013recursive}, and \textit{News20}~\cite{lang1995newsweeder} are text datasets with 10,000 samples. For the \textit{CIFAR10} and text datasets, we employ the ResNet50~\cite{he2016deep} and Sentence Bert~\cite{reimers2019sentence} as embedding for the KNN classifier, respectively. Features/embeddings of these nine datasets are from 14 to 2,048. The detailed characteristics of these datasets can be found in Appendix~\ref{app:dataset}. For the baseline methods, we choose the KNN-Shapley~\cite{jia2018efficient}, KNN-Shapley-JW~\cite{wang2023note}, and TKNN-Shapley~\cite{wang2023threshold} with default setting $K$$=$$10$ and $\tau$$=$$-0.5$. Besides, we set $T$$=$$N$$-$$2K$ in our CKNN-Shapley.

\begin{figure*}[!t]
        \centering
        \subfigure{
            \includegraphics[width=0.31\textwidth]{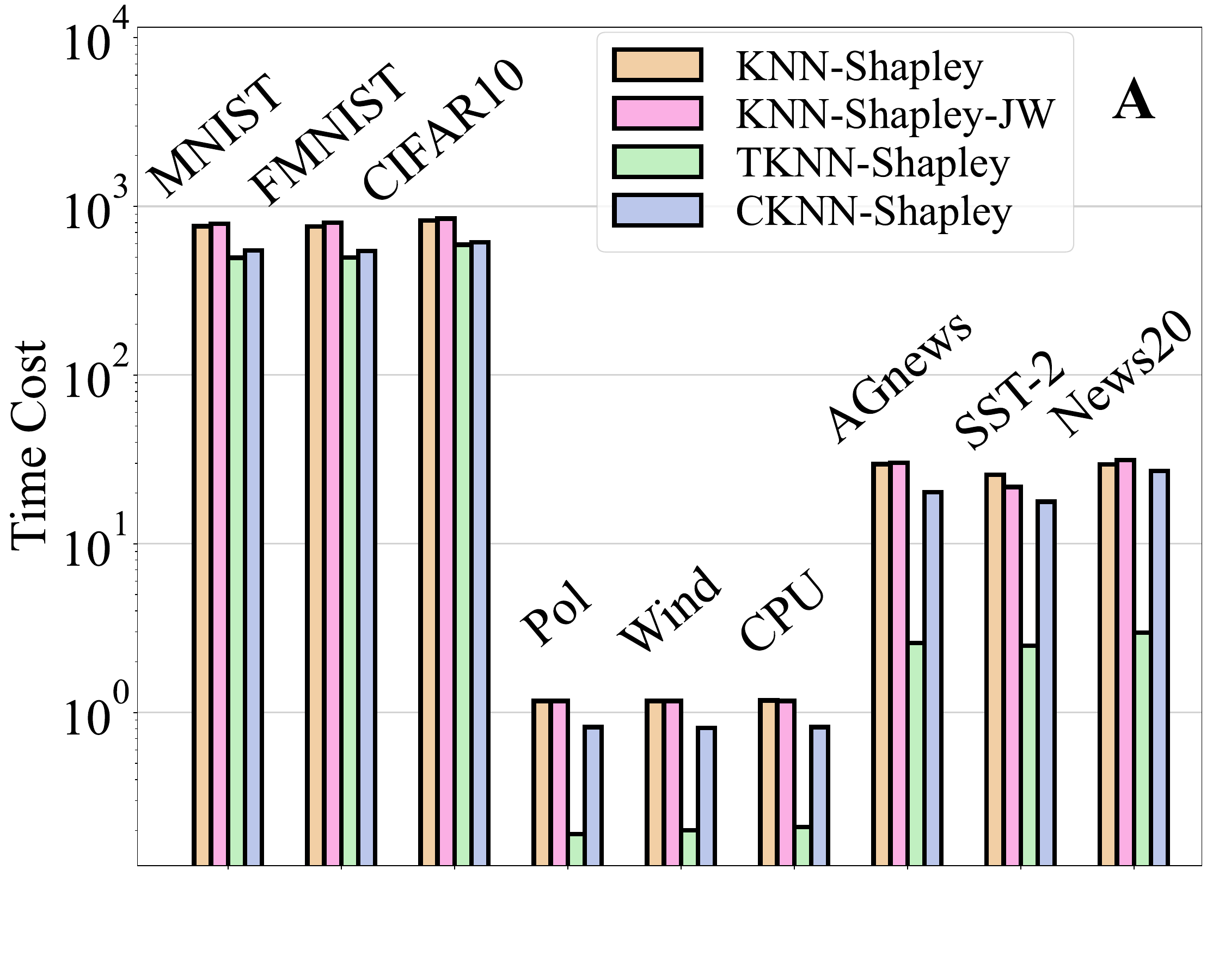} 
            }
        \subfigure{
            \includegraphics[width=0.31\textwidth]{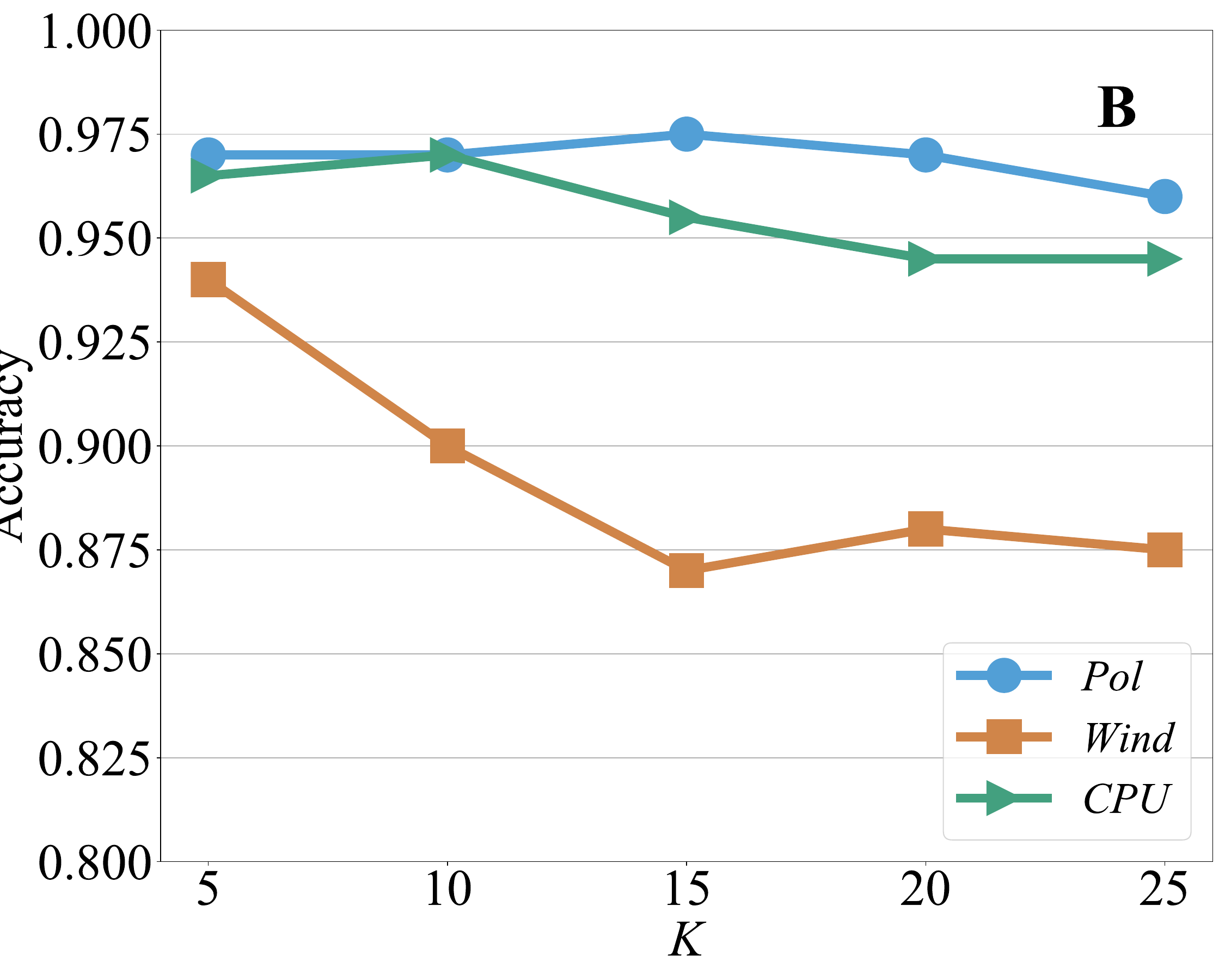} 
            }
        \subfigure{
            \includegraphics[width=0.31\textwidth]{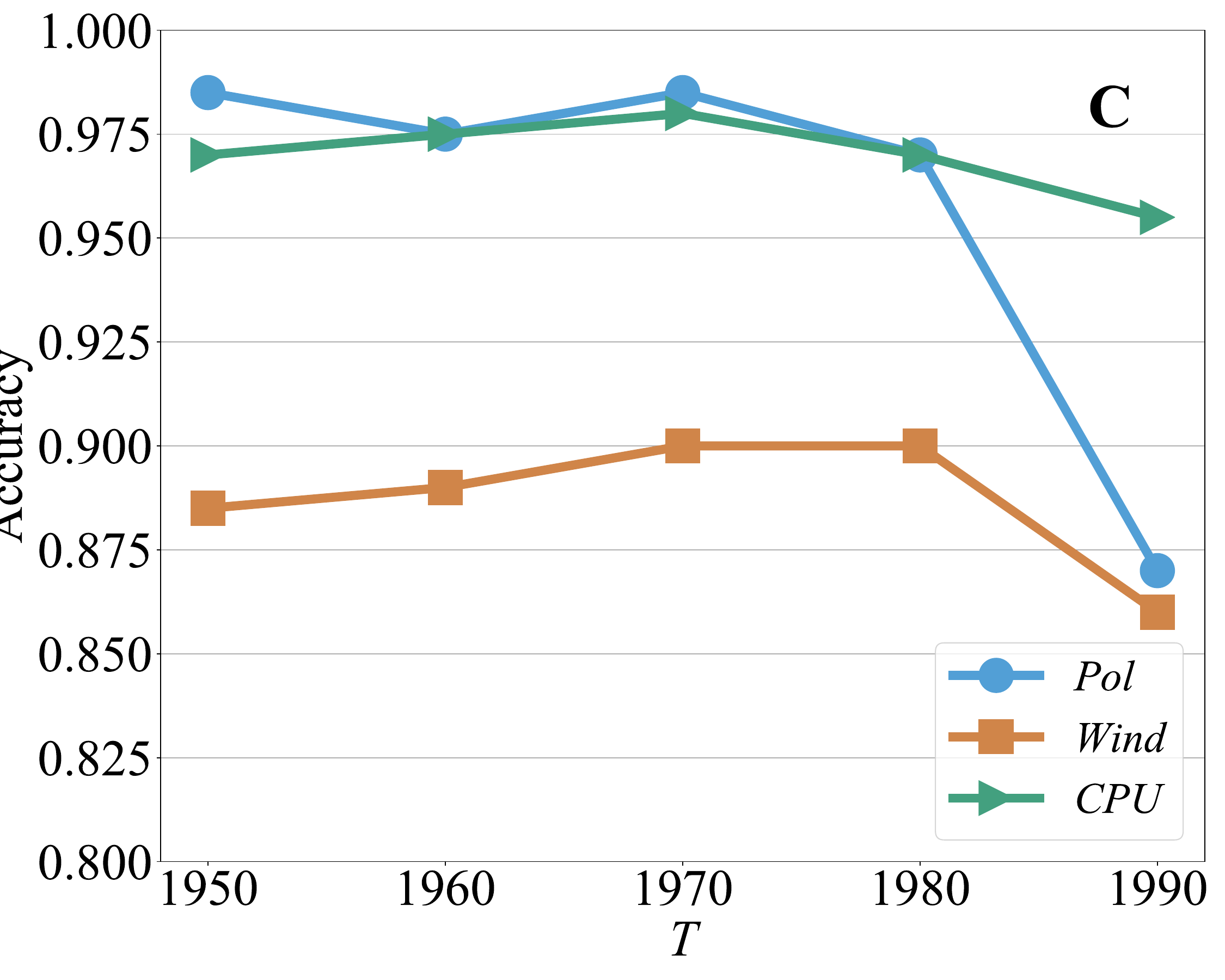} 
            }
        \vspace{-6mm}
        \caption{Execution time and parameter analysis. \textbf{A} shows the execution time by second in the logarithm of three KNN-Shapley-based data valuation approaches; \textbf{B} and \textbf{C} display the classification performance trend of our CKNN-Shapley with different values of $K$ and $T$.}
        \label{fig:time_k_t}
        \vspace{-4mm}
    \end{figure*}

In addition to the conventional predictive accuracy for classification, we introduce two metrics to assess valuation inflation: the threshold for distinguishing detrimental samples from beneficial ones and the misidentification ratio of detrimental samples. This approach is inspired by the illustration in Figure~\ref{fig:comparison}. The dataset is divided into 100 bins, and the goal is to find the intersection of the red and green lines to determine the threshold for distinguishing detrimental samples. The misidentification ratio is then calculated as the proportion of samples in the green region to all detrimental samples. The entire training dataset is segmented into 100 equal-size bins based on the ascending order of data valuation, denoted as $\nu_i$ for the data valuation of the $i$-th bin. Let $p_i$ represent the KNN classifier's performance on the training set without samples in the $i$-th bin, and $p_0$ represent the performance on the entire training set. We define the threshold $t$ for distinguishing detrimental samples from beneficial ones and the misidentification ratio $r$ of detrimental samples as follows:
\begin{equation}\label{eq:metric}
    t=\nu_{j^*}\ \textup{and} \ r= (j^* - i^*)/ j^*,
\end{equation}
where $j^*= \min_{j}\{p_j<p_0\ \&\ p_{j+1}<p_0\}$ and $i^*$ is the index of $i^*$-bin with $\nu_{i^*}=0$. 

All experiments were conducted on a workstation with an AMD Ryzen Threadripper PRO 5965WX CPU and x86\_64 architecture with 128 GB memory.

\begin{table*}[t]
\centering
\vspace{-4mm}
\caption{Classification performance of KNN-Shapley-based methods with fixed numbers of removed samples. $\#$, $+$, and $*$ denote KNN's performance trained on the training set excluding samples with the smallest 10\%, 20\%, and 30\% data valuations, respectively.}\label{tab:removefix}\vspace{1mm}
\resizebox{0.98\textwidth}{!}{%
\begin{tabular}{p{3.5cm}|*{9}{>{\centering\arraybackslash}p{1.2cm}}|>{\centering\arraybackslash}p{1.2cm}}
\toprule
Method$\backslash$Datasets  & \textit{MNIST} & \textit{FMNIST} & \textit{CIFAR10} & \textit{Pol} & \textit{Wind} & \textit{CPU} & \textit{AGnews} & \textit{SST-2} & \textit{News20} & Avg. \\ 
 \midrule
Vanilla KNN & 0.9630 & 0.8444 & 0.5956 & 0.9400 & 0.8700 & 0.9200 & 0.9060 & 0.7270 & 0.6920 & 0.8287\\
 \midrule
% LOO$^\#$ & - & \\ 
KNN-Shapley~\cite{jia2018efficient}$^\#$ & 0.9702 & 0.8590 & 0.6270 & \textbf{0.9750} & 0.8850 & 0.9550 & 0.9250 & 0.8016 & 0.7290 & 0.8585\\
KNN-Shapley-JW~\cite{wang2023note}$^\#$ & 0.9702 & 0.8590 & 0.6270 & 0.9700 & 0.8850 & 0.9550 & 0.9250 & 0.8016 & 0.7280 & 0.8579\\ 
TKNN-Shapley~\cite{wang2023threshold}$^\#$ & 0.9516 & 0.8368 & 0.5670  & 0.9450 & 0.8450 & 0.9200 & 0.9190 & 0.7878 & 0.7160 & 0.8320\\ 
CKNN-Shapley (Ours)$^\#$ & \textbf{0.9812} & \textbf{0.8822} & \textbf{0.6566} & 0.9650 & \textbf{0.8950} & \textbf{0.9700} & \textbf{0.9370} & \textbf{0.8108} & \textbf{0.7450}  & \textbf{0.8714}\\ 
 \midrule
% LOO$^+$ & - & \\ 
KNN-Shapley~\cite{jia2018efficient}$^+$ & 0.9714 & 0.8548 & 0.6396 & 0.9650 & 0.8650 & 0.9550 & 0.9250 & 0.8314 & 0.7460 & 0.8615\\ 
KNN-Shapley-JW~\cite{wang2023note}$^+$ & 0.9714 & 0.8548 & 0.6396 & 0.9650 & 0.8650 & 0.9550 & 0.9250 & 0.8303 & 0.7470 & 0.8615\\ 
TKNN-Shapley~\cite{wang2023threshold}$^+$ & 0.8744 & 0.7724 & 0.5292 & 0.8850 & 0.8350 & 0.9300 & 0.9170 & 0.8005 & 0.7190 & 0.8069\\ 
CKNN-Shapley (Ours)$^+$ & \textbf{0.9814} & \textbf{0.8874} & \textbf{0.6990} & \textbf{0.9700} & \textbf{0.9100} & \textbf{0.9800} & \textbf{0.9450} & \textbf{0.8830} & \textbf{0.7800} & \textbf{0.8929} \\ 
 \midrule
% LOO$^+$ & - & \\ 
KNN-Shapley~\cite{jia2018efficient}$^*$& 0.9726 & 0.8536 & 0.6482 & 0.9650 & 0.8600 & 0.9450 & 0.9250 & 0.8452 & 0.7620 & 0.8641\\ 
KNN-Shapley-JW~\cite{wang2023note}$^*$& 0.9726 & 0.8536 & 0.6482 & \textbf{0.9700} & 0.8600 & 0.9550 & 0.9250 & 0.8463 & 0.7600 & 0.8656\\ 
TKNN-Shapley~\cite{wang2023threshold}$^*$ & 0.8378 & 0.7130 & 0.4716 & 0.8400 & 0.8300 & 0.9100 & 0.9180 & 0.7982 & 0.7100 & 0.7810\\ 
CKNN-Shapley (Ours)$^*$ & \textbf{0.9814} & \textbf{0.8882} & \textbf{0.7174} & \textbf{0.9700} & \textbf{0.9100} & \textbf{0.9750} & \textbf{0.9450} & \textbf{0.8968} & \textbf{0.7990} & \textbf{0.8981} \\ 
\bottomrule
\end{tabular}\vspace{-8mm}
}
\label{tab:102030percent}
\end{table*}

\textbf{Algorithmic Performance}. We present the algorithmic comparison of different KNN-Shapley-based methods, focusing on value inflation and classification performance. Table~\ref{tab:index} displays the thresholds for distinguishing detrimental from beneficial samples and the misidentification ratio of detrimental samples. KNN-Shapley, KNN-Shapley-JW, and TKNN-Shapley exhibit thresholds far from zero, spanning a wide range across different datasets from 0.1632 to 6.4458, from 0.8203 to 4.6543, and from 0.7143 to 5.6960, respectively. This variability complicates the interpretation of their data valuation. In contrast, our CKNN-Shapley consistently achieves thresholds close to zero, enhancing the meaningfulness of data valuation in practical applications. The misidentification ratios of detrimental samples provide additional insights into these thresholds in terms of sample proportions. KNN-Shapley exhibits substantial misidentification ratios, surpassing 50\% on 6 out of 9 datasets. KNN-Shapley-JW also surpasses 50\% on 3 out of 9 datasets. TKNN-Shapley shows slightly better performance in terms of the threshold but still exceeds 41\% inflation on \textit{CPU}, while CKNN-Shapley with calibrated thresholds, averages only around 10\% inflation of detrimental sets and achieves no inflation on \textit{Pol}, \textit{Wind}, and \textit{CPU}.

In addition to evaluating value inflation, we analyze the impact of inflation on classification performance by removing samples with negative valuation and using weighted KNN. Table~\ref{tab:remove} compares the results. KNN-Shapley consistently improves performance on all datasets by removing samples with negative KNN-Shapley values, outperforming vanilla KNN with all training samples. However, TKNN-Shapley, designed for membership protection, does not yield desirable performance, possibly due to default parameter settings. In contrast, CKNN-Shapley mitigates the negative effects of inflation on detrimental sets, providing additional performance boosts on almost every dataset, except \textit{Pol} where both KNN-Shapley and CKNN-Shapley achieve the same performance. Notably, on \textit{CIFAR10} and \textit{SST-2}, CKNN-Shapley exhibits over 7\% and 8\% performance improvements over KNN-Shapley, addressing the 49\% and 56\% inflation in misidentification ratios observed in KNN-Shapley on these datasets. In addition to detrimental set inflation, Figure~\ref{fig:comparison} reveals inflation in the beneficial set. We further conduct experiments with a weighted KNN classifier, using weights derived from data valuations, and observe similar phenomena. CKNN-Shapley achieves significant improvements with calibrated data valuations, underscoring the need to address value inflation issues. To address the potential impact of the number of removed samples, we further investigate the setting of removing a fixed amount of samples, as shown in Table~\ref{tab:removefix}. Our CKNN-Shapley consistently demonstrates superior performance over various KNN-Shapley-based methods. We extend our analysis by conducting additional experiments that explore misidentified samples and demonstrate the generalization of our CKNN-Shapley approach to other classifiers. Detailed results are presented in Appendix~\ref{app:additional}.

Figure~\ref{fig:time_k_t}\textbf{A} shows the execution time of four methods, where TKNN-Shapley has the fastest speed due to its linear time complexity. For KNN-Shalpley, KNN-Shapley-JW, and our CKNN-Shapley, all of them have the $\mathcal{O}(N\log N)$; CKNN-Shapley is faster since $T/N$ percentage valuations are directly assigned to zero without the recursive calculation. Figure~\ref{fig:time_k_t}\textbf{B} and \textbf{C }display the classification performance trend of our CKNN-Shapley with different values of $K$ and $T$, where a small $K$ and large $T$ but not close to $N$ are preferred. A small $K$ excludes the samples from different categories. Besides, a large $T$ enforces the similar to the original one, while maintaining the diversity of selected training subset. We posit that the choice of $T$ is contingent upon the dataset characteristics. In Appendix~\ref{app:additional}, we explore various settings of $T$ and ascertain that $N$$-$$2K$ serves as a suitable setting.

 \begin{figure*}[t]
        \centering
        \subfigure{
            \includegraphics[width=0.31\textwidth]{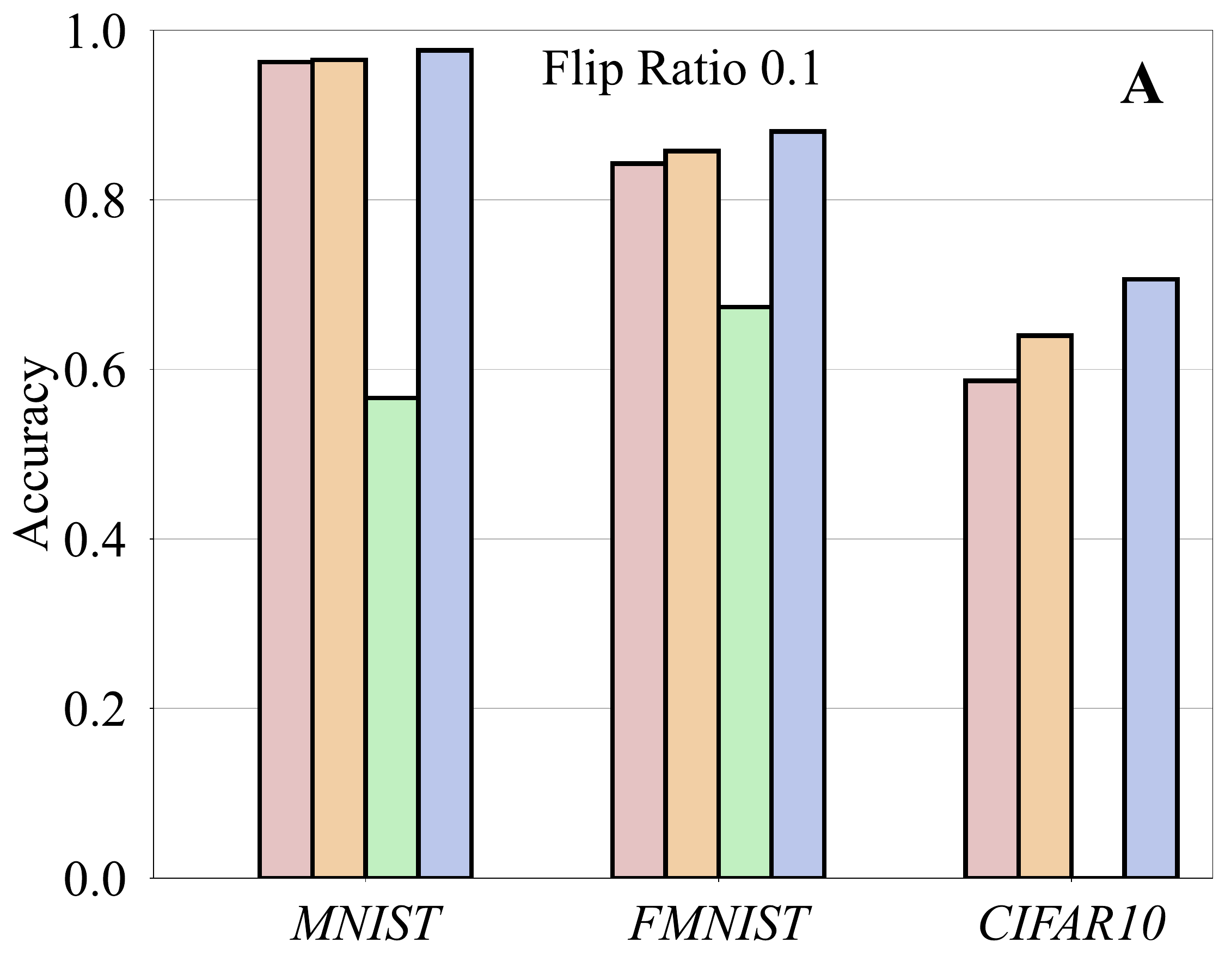} 
            } \vspace{-2mm}
        \subfigure{
            \includegraphics[width=0.31\textwidth]{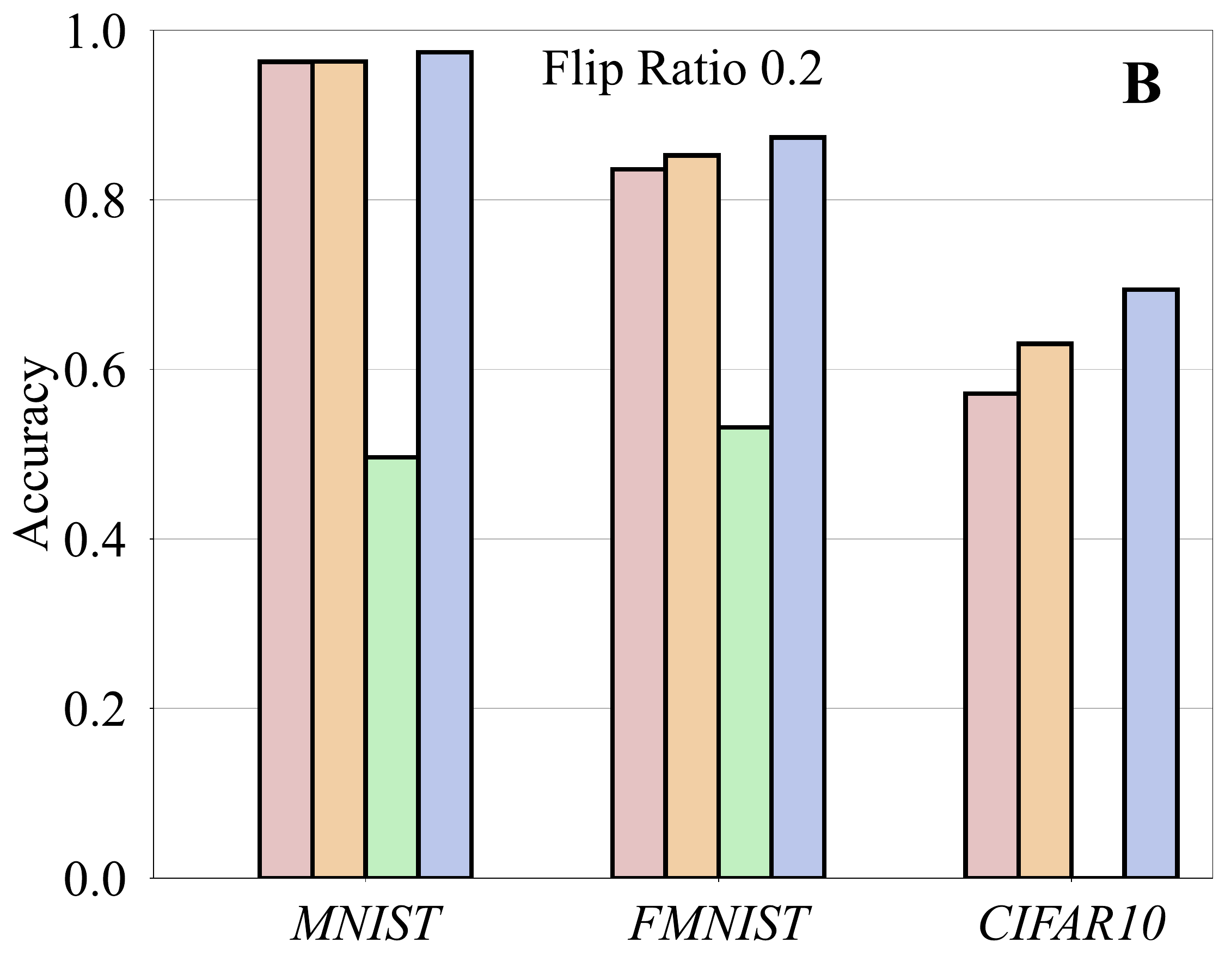} 
            } \vspace{-2mm}
        \subfigure{
            \includegraphics[width=0.31\textwidth]{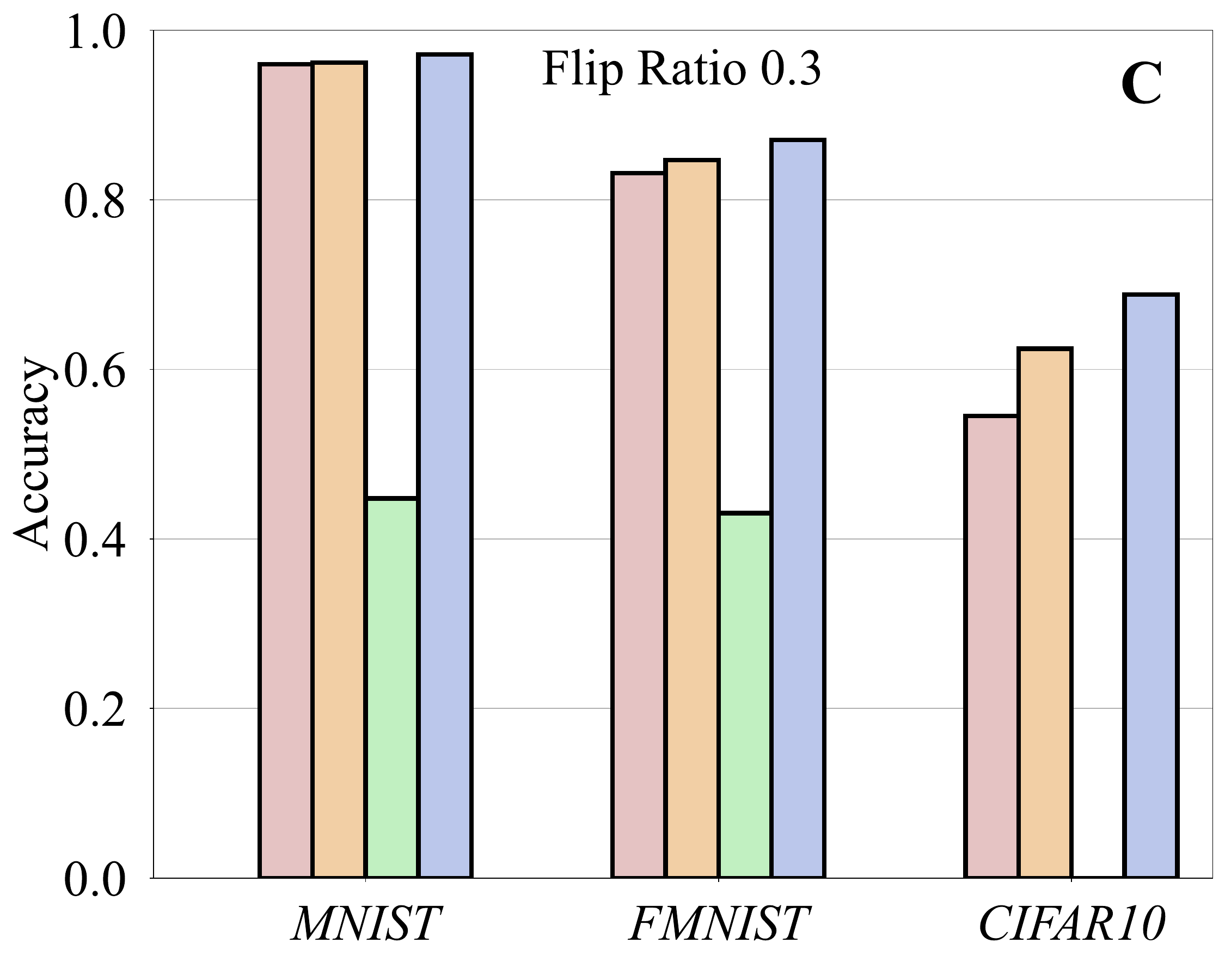} 
            } \vspace{-2mm}
        \subfigure{
            \includegraphics[scale=0.13]{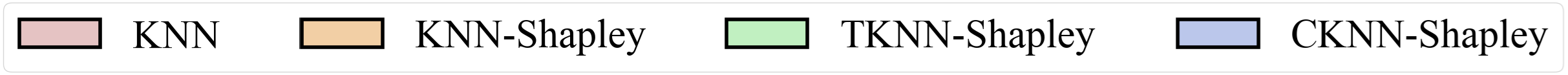} \vspace{-2mm}
            }
        \vspace{-2mm}
        \caption{The classification performance of KNN on datasets \textit{MNIST}, \textit{FMNIST}, and \textit{CIFAR10} varies with different training sets and flip ratios. The standard KNN utilizes the full training set, including mislabeled data, whereas KNN-Shapley-based methods start by excluding samples having negative Shapley values from the training set, and then apply the KNN classifier.}
        \label{fig:mislabel}
        \vspace{-4mm}
    \end{figure*}

    \begin{figure*}[t]
        \centering
        \subfigure{
            \includegraphics[width=0.26\textwidth]{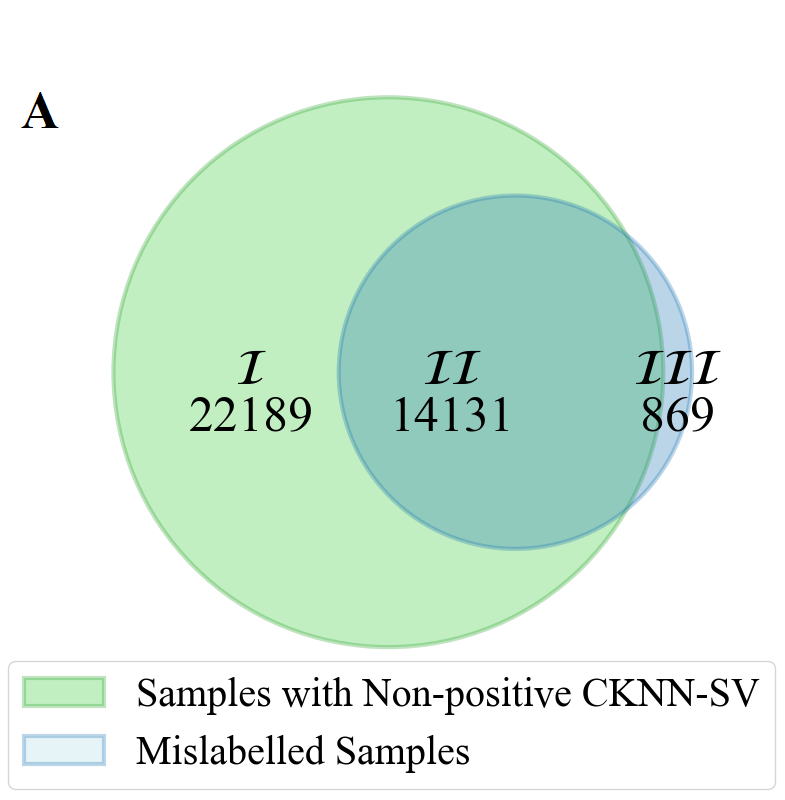} 
            }
        \subfigure{
            \includegraphics[width=0.32\textwidth]{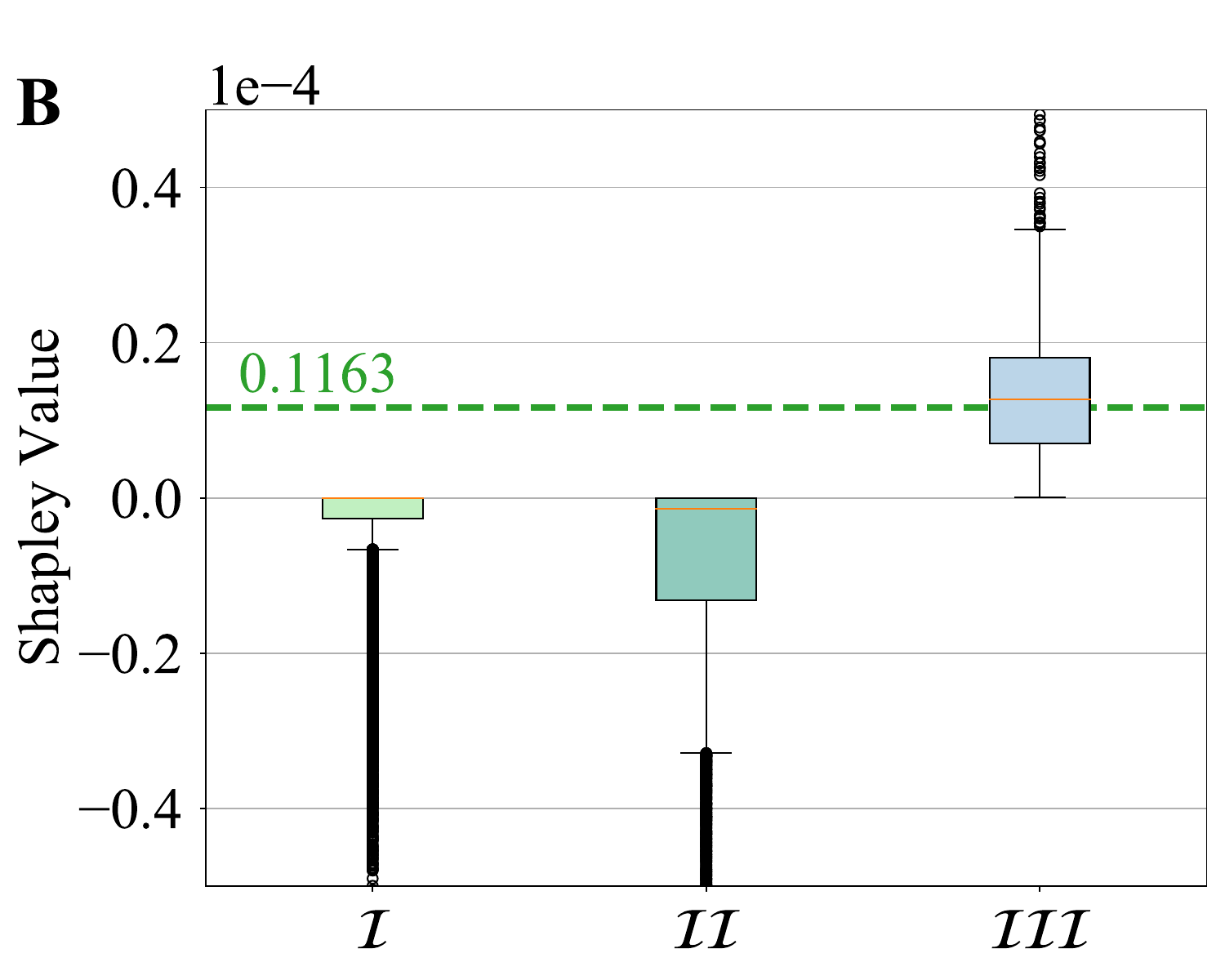} 
            }
        \subfigure{
            \includegraphics[width=0.36\textwidth]{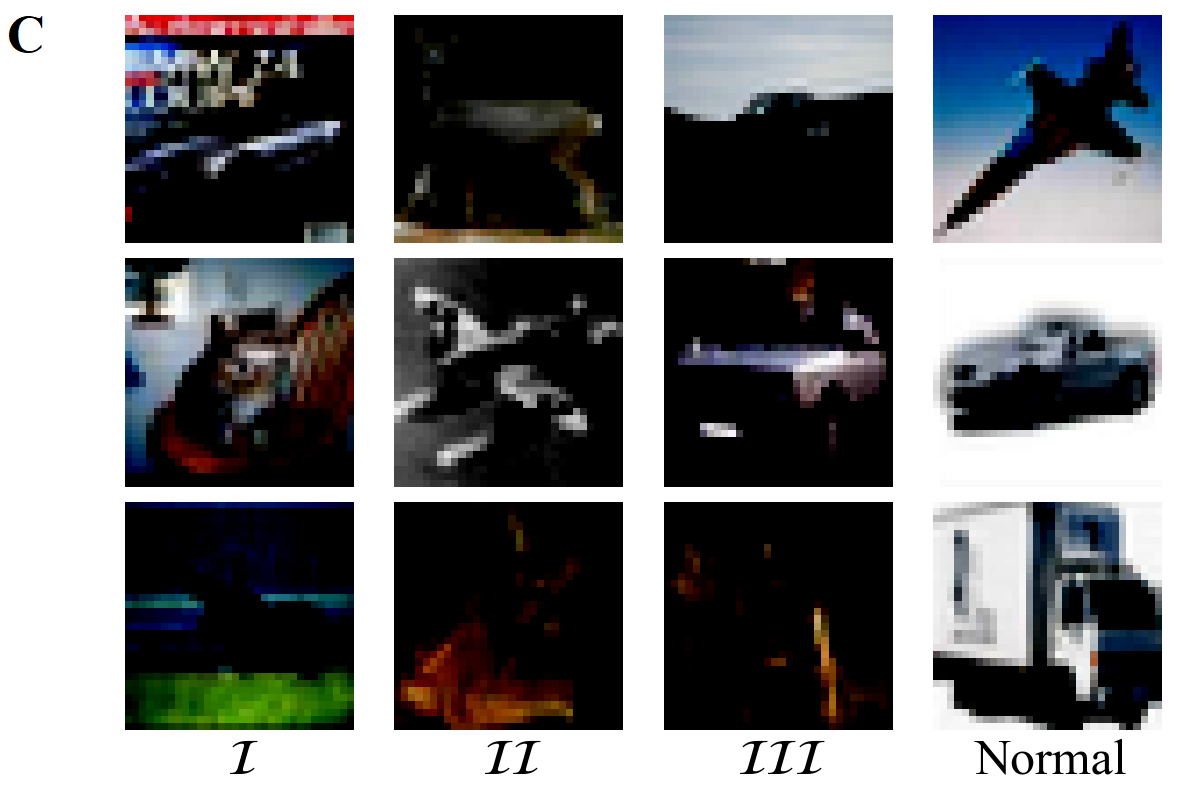} 
            }
        \vspace{-4mm}
        \caption{In-depth exploration of CKNN-Shapley on \textit{CIFAR10} with 0.3 flip ratio. \textbf{A} depicts the sizes of detrimental and mislabelled samples, where $\mathcal{I}$ denotes the set of samples with non-positive Shapley values but not mislabeled, $\mathcal{II}$ presents the set of joint detrimental and mislabeled samples, and $\mathcal{III}$ is the set of mislabeled samples with positive Shapley values. \textbf{B} shows the value distribution of these three sets. \textbf{C} displays the visual examples of detrimental or mislabeled samples and normal samples.}
        \label{fig:indepth}
        \vspace{-4mm}
    \end{figure*}

\section{Applications Beyond Conventional Classification}
We extend the conventional classification in the previous section by conducting extensive experiments for various practical scenarios, and demonstrate the effectiveness of our proposed CKNN-Shapley in resisting mislabeled data, mitigating distribution shift, and identifying beneficial samples.\footnote{We omit KNN-Shapley-JW in this section due to its similar performance with KNN-Shapley.}

\textbf{Learning with Mislabeled Data}. 
We manually simulate mislabeled data by randomly flipping its original label into another category and conduct experiments on \textit{MNIST}, \textit{FMNIST}, and \textit{CIFAR10} with different flip ratios. In Figure~\ref{fig:mislabel}, we observe the KNN performance with different training sets, where vanilla KNN runs on the complete training set with mislabeled data, and KNN-Shapley-based methods first remove samples with negative Shapley values from the training set before training the KNN classifier. In general, KNN-Shapley and CKNN-Shapley effectively resist the negative impact of mislabeled data, outperforming vanilla KNN. Unfortunately, TKNN-Shapley performs poorly, significantly worse than vanilla KNN, and provides no meaningful results on \textit{CIFAR10}. Due to CKNN-Shapley addressing the value inflation of KNN-Shapley, its performance is further improved and consistently achieves the best results across all three datasets. Notably, CKNN-Shapley maintains similar performance across different flip ratios, demonstrating its resilience to mislabeled data.

Furthermore, we delve into the details of CKNN-Shapley on \textit{CIFAR10} with a 0.3 flip ratio, as shown in Figure~\ref{fig:indepth}. In Figure~\ref{fig:indepth}\textbf{A}, we examine the Shapley values of mislabeled samples, separating the detrimental and mislabeled sets into three categories: $\mathcal{I}$ denotes the set of samples with non-positive Shapley values but not mislabeled, $\mathcal{II}$ represents the set of joint detrimental and mislabeled samples, and $\mathcal{III}$ is the set of mislabeled samples with positive Shapley \mbox{values}. We observe that the majority of mislabeled samples are associated with negative Shapley values, indicating their detrimental nature. Figure~\ref{fig:indepth}\textbf{B} presents the value distribution of the three sets, revealing a small portion of mislabeled data with very small positive values. From Table~\ref{tab:index}, we find that the threshold for distinguishing detrimental and beneficial samples on \textit{CIFAR10} is around $0.11\times$$e-$$4$. Several visual examples in Figure~\ref{fig:indepth}\textbf{C} illustrate the \mbox{significant} distinction between samples $\mathcal{I}, \mathcal{II}$, and $\mathcal{III}$ and normal samples. Samples in $\mathcal{I}, \mathcal{II}$, and $\mathcal{III}$ have a dark background, making them difficult to recognize.

\begin{figure*}[t]
        \centering
        \subfigure{
            \includegraphics[width=0.31\textwidth]{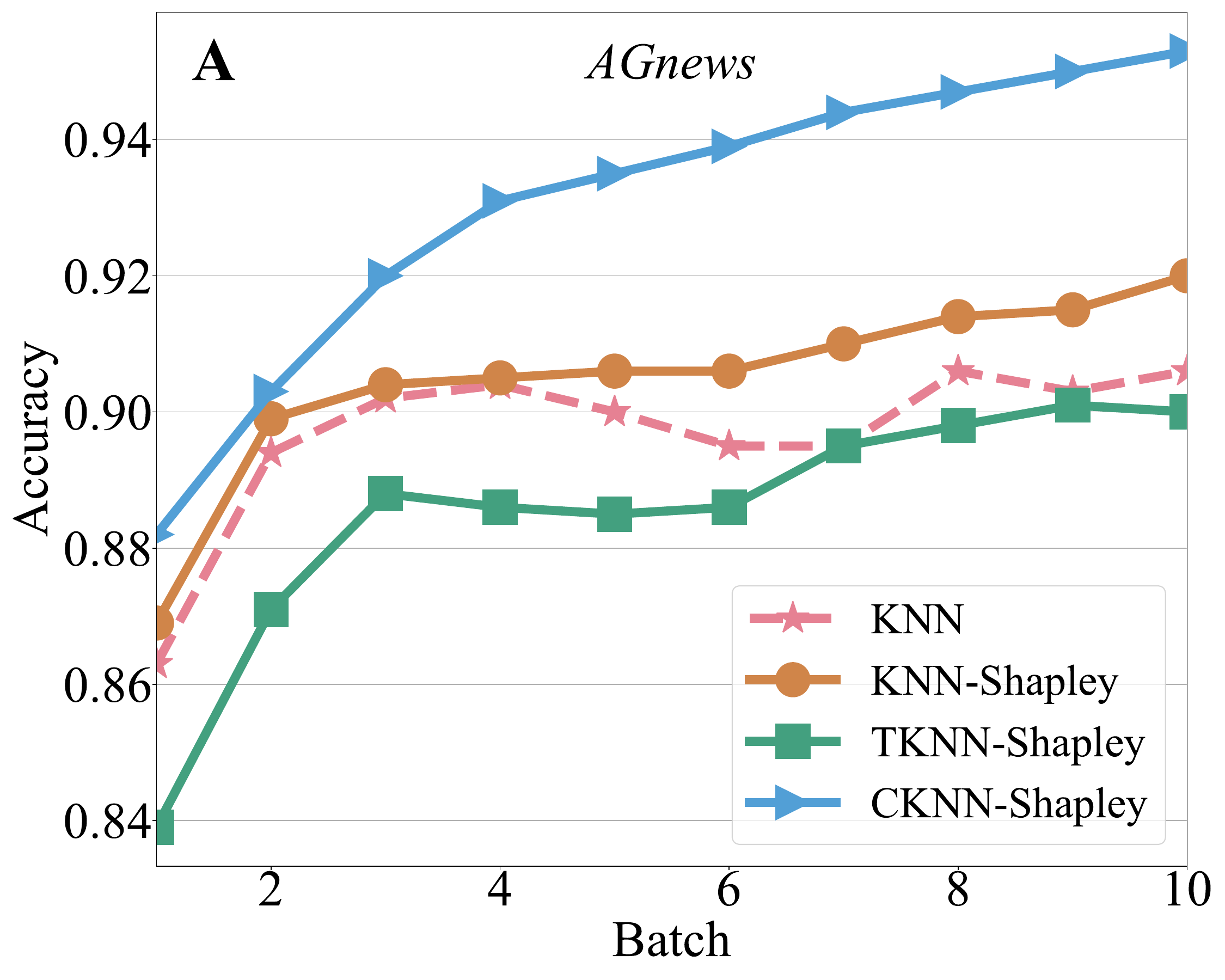} 
            }\vspace{-1.5mm}
        \subfigure{
            \includegraphics[width=0.31\textwidth]{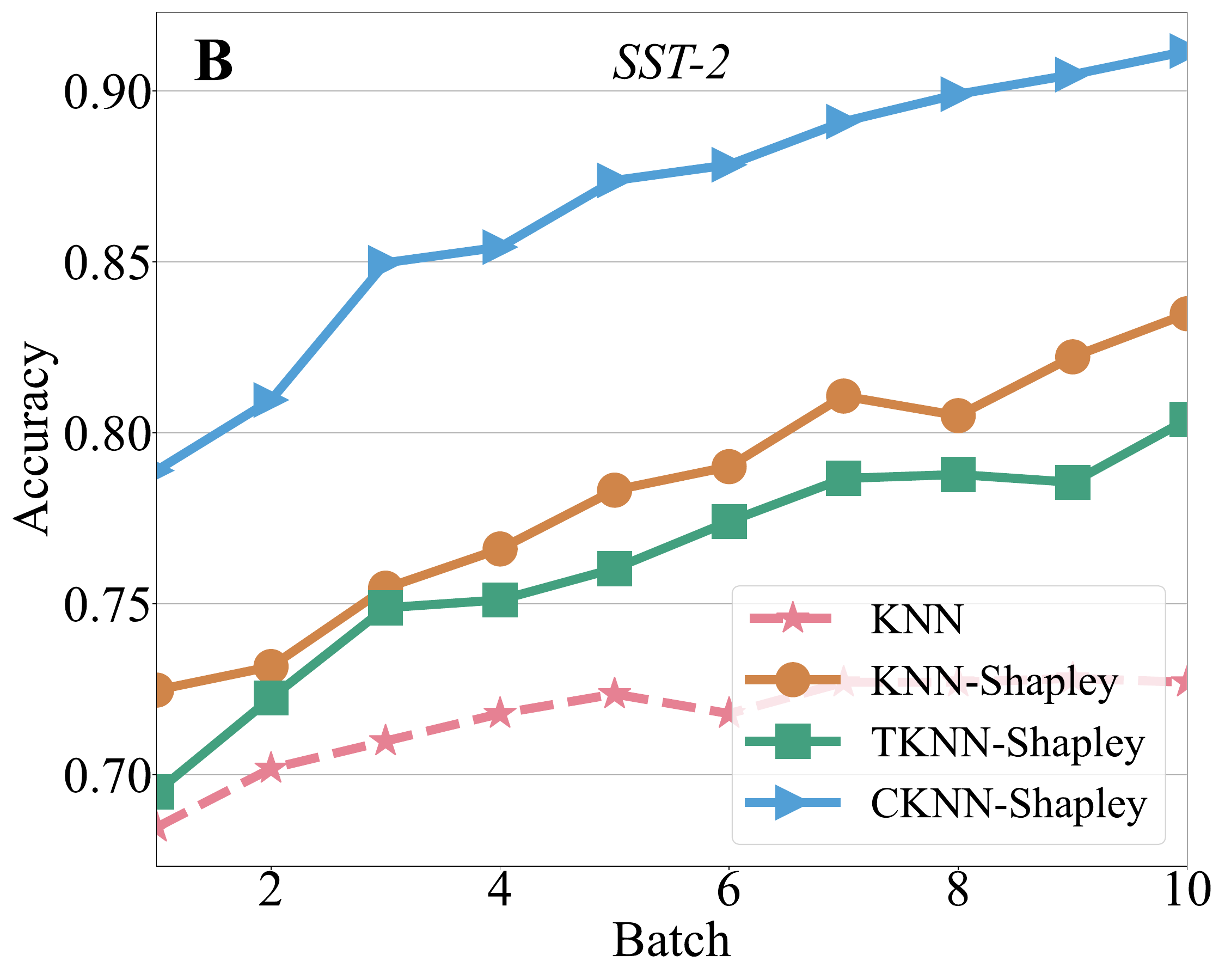} 
            }\vspace{-1.5mm}
        \subfigure{
            \includegraphics[width=0.31\textwidth]{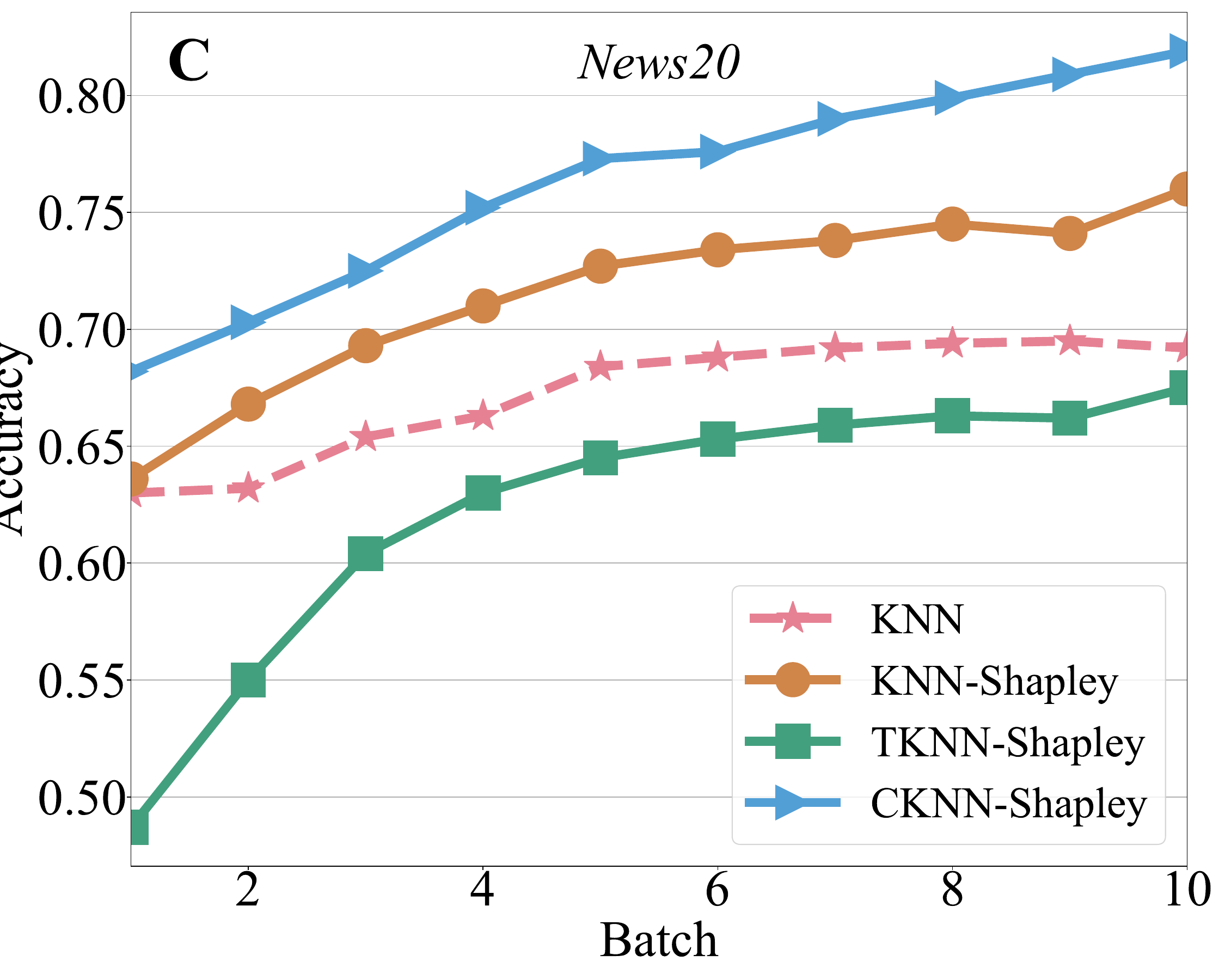} 
            }\vspace{-1.5mm}

        \subfigure{
            \includegraphics[width=0.31\textwidth]{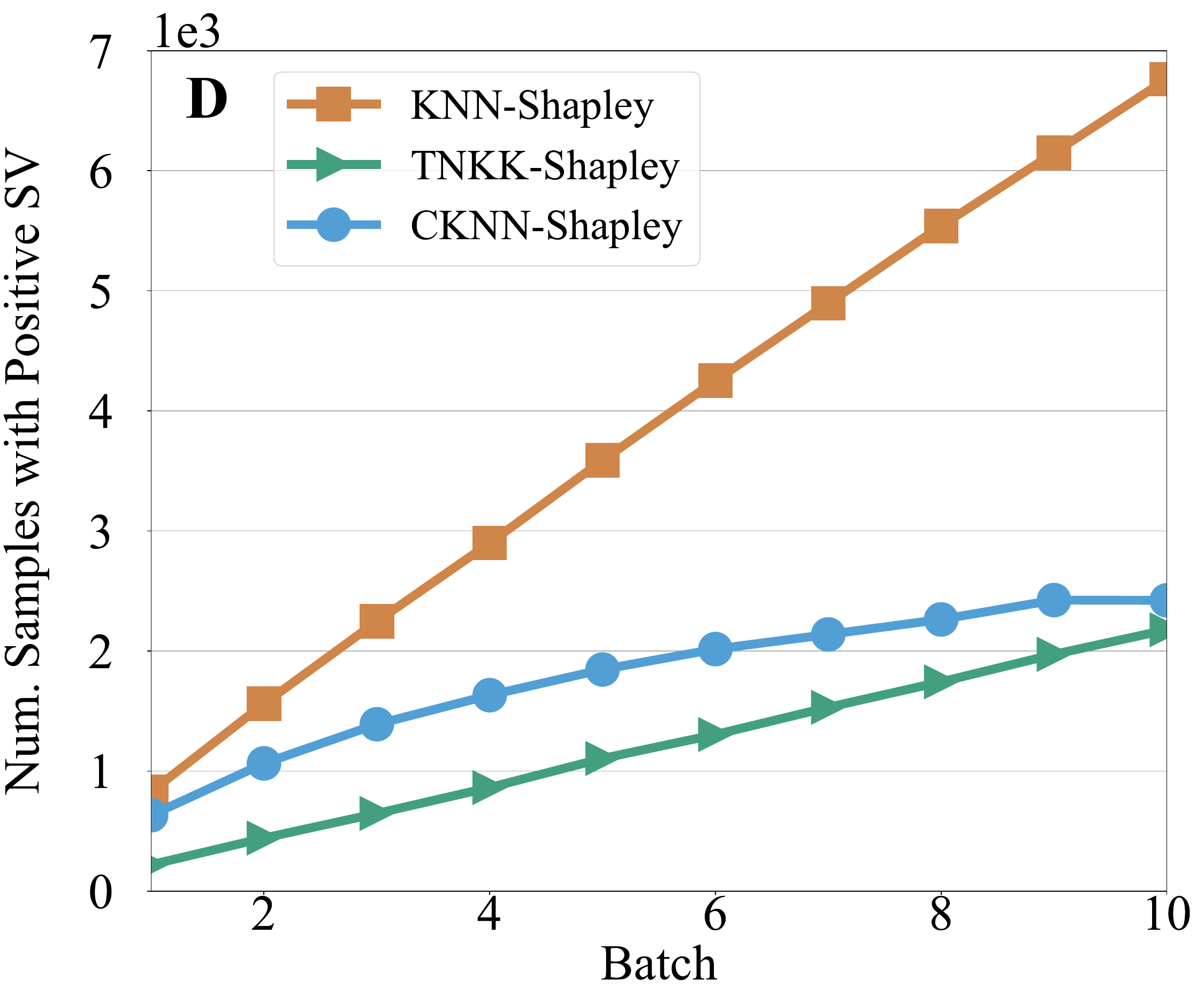} 
            }
        \subfigure{
            \includegraphics[width=0.31\textwidth]{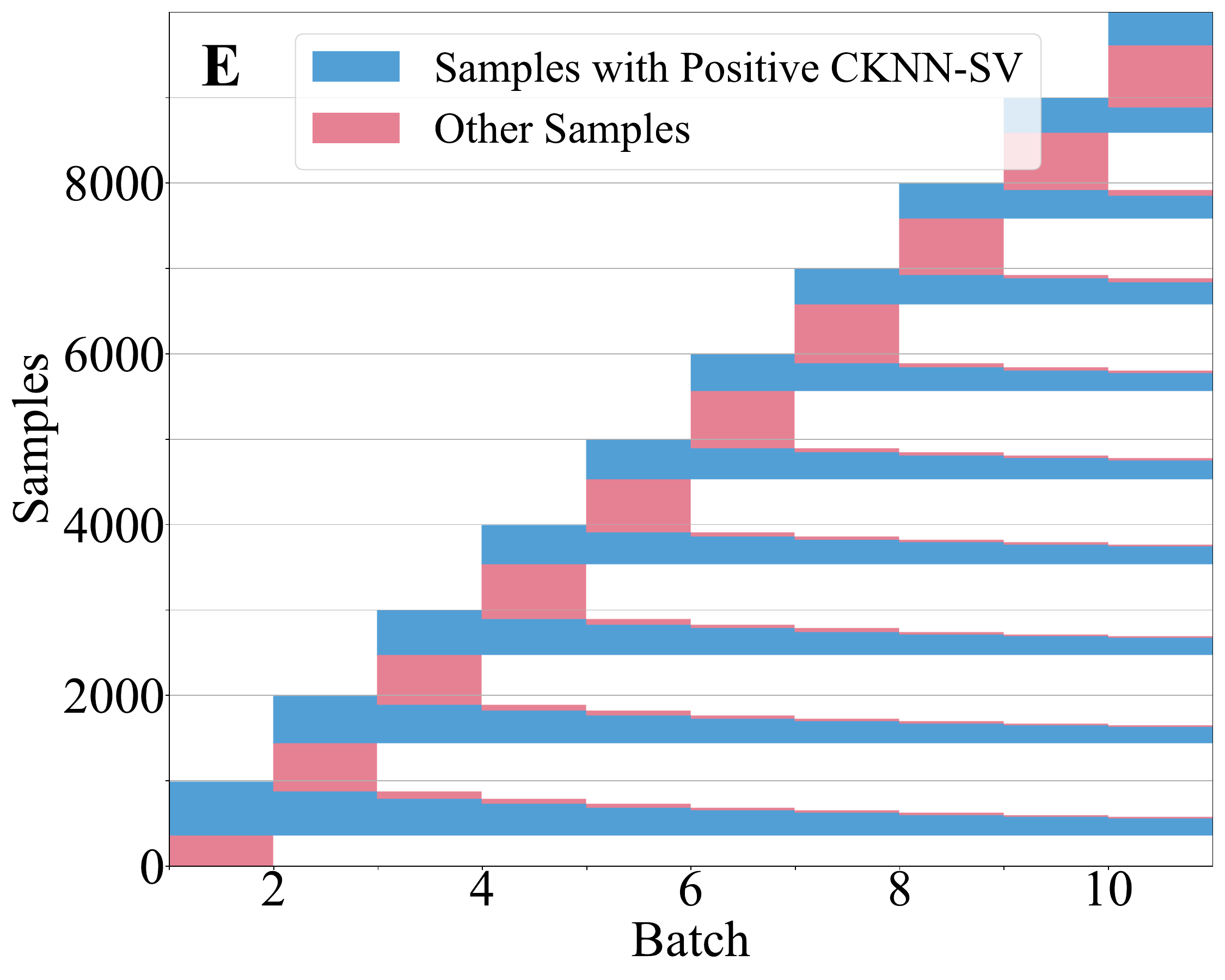} 
            }
        \subfigure{
            \includegraphics[width=0.31\textwidth]{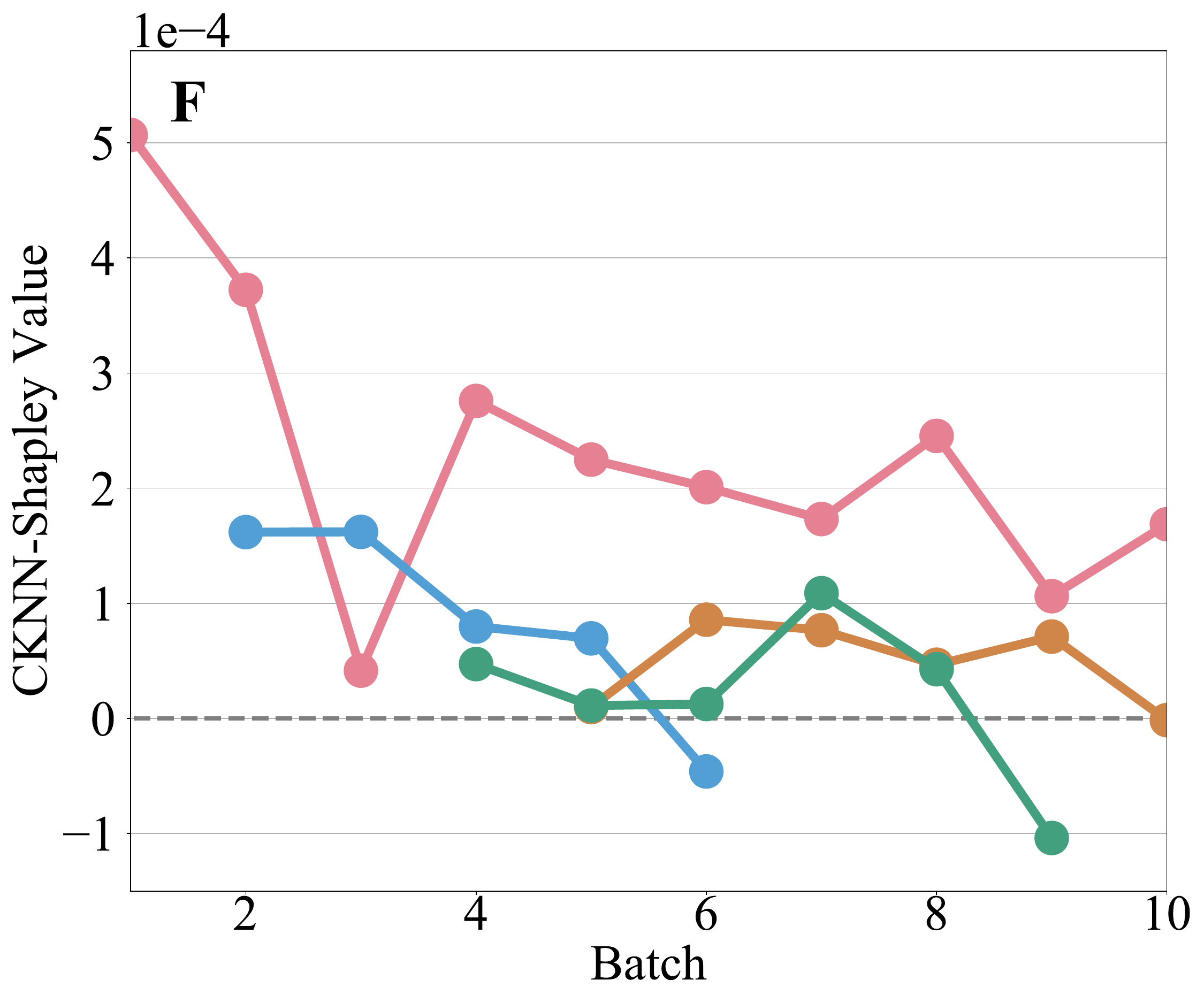} 
            }
        \vspace{-4mm}
        \caption{Performance of online learning. \textbf{A-C} depict the online learning performance of KNN-Shapley-based data valuation approaches by removing samples with negative Shapley values on the three text datasets across different batches, where the dashed lines present the performance of vanilla KNN without removing any samples. \textbf{D} displays the remaining training samples in each batch, combining samples from the last batch and new arrivals in the current batch by removing detrimental samples. \textbf{E} provides details of involved training samples for CKNN-Shapley on \textit{News20} across different batches, followed by the Shapley value changes of some representative samples in \textbf{F}.}
        \label{fig:online}
        \vspace{-4mm}
    \end{figure*}

\textbf{Online Learning}. 
In addition to static datasets, we further evaluate our approach in the context of online stream data. Specifically, we use three text datasets—\textit{AGnews}, \textit{SST-2}, and \textit{News20}—in this experiment. We segment these datasets into 10 equal-sized batches and gradually feed them into the KNN classifier to simulate stream data. KNN-Shapley-based data valuation approaches are employed to remove detrimental samples with negative Shapley values. Figures~\ref{fig:online}\textbf{A-C} depict the online learning performance of three KNN-Shapley-based data valuation approaches on the three text datasets across different batches. CKNN-Shapley consistently outperforms other methods with large margins, resolving the value inflation issue and showing more effectiveness in identifying detrimental samples in the stream context. Note that an increasing margin of CKNN-Shapley over others can be observed on \textit{AGnews}. Figure~\ref{fig:online}\textbf{D} displays the remaining training samples in each batch, combining samples from the last batch and new arrivals in the current batch by removing detrimental samples. KNN-Shapley involves more training samples compared to others, indicating severe inflation issues. In contrast, CKNN-Shapley removes a significant portion of detrimental samples, enhancing learning performance with fewer training samples. Figure~\ref{fig:online}\textbf{E} provides details of involved training samples for CKNN-Shapley on \textit{News20} across different batches. The blue color denotes a sample included in a batch, while the pink color denotes its removal from the current and subsequent batches. Additionally, Figure~\ref{fig:online}\textbf{F} shows the Shapley value change for representative samples. The pink line represents a sample included in all batches, while other lines represent samples initially associated with positive Shapley values in the initial batches, which later become negative and are subsequently removed.

\begin{figure*}[!t]
        \centering
        \subfigure{
            \includegraphics[width=0.31\textwidth]{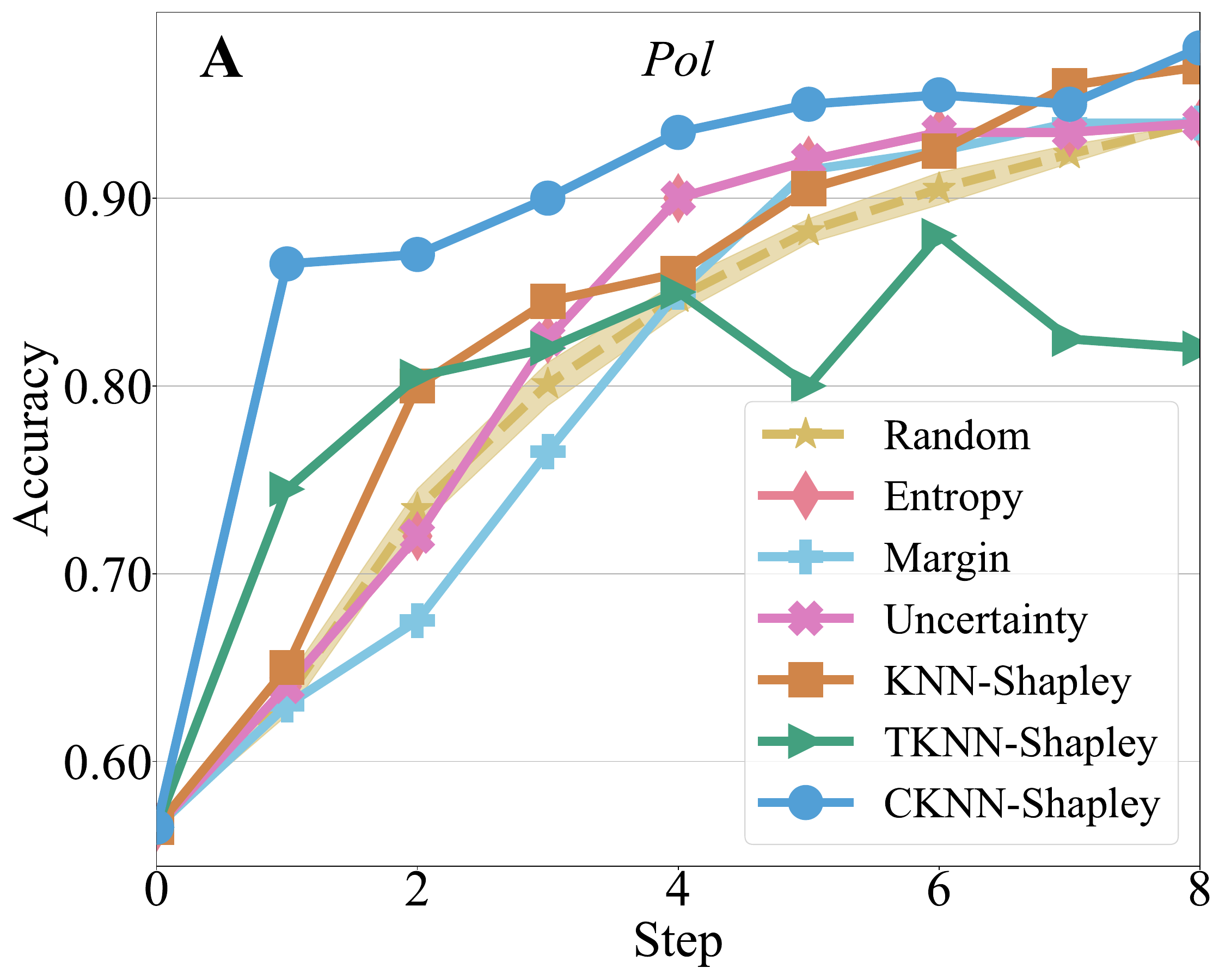} 
            }
        \subfigure{
            \includegraphics[width=0.31\textwidth]{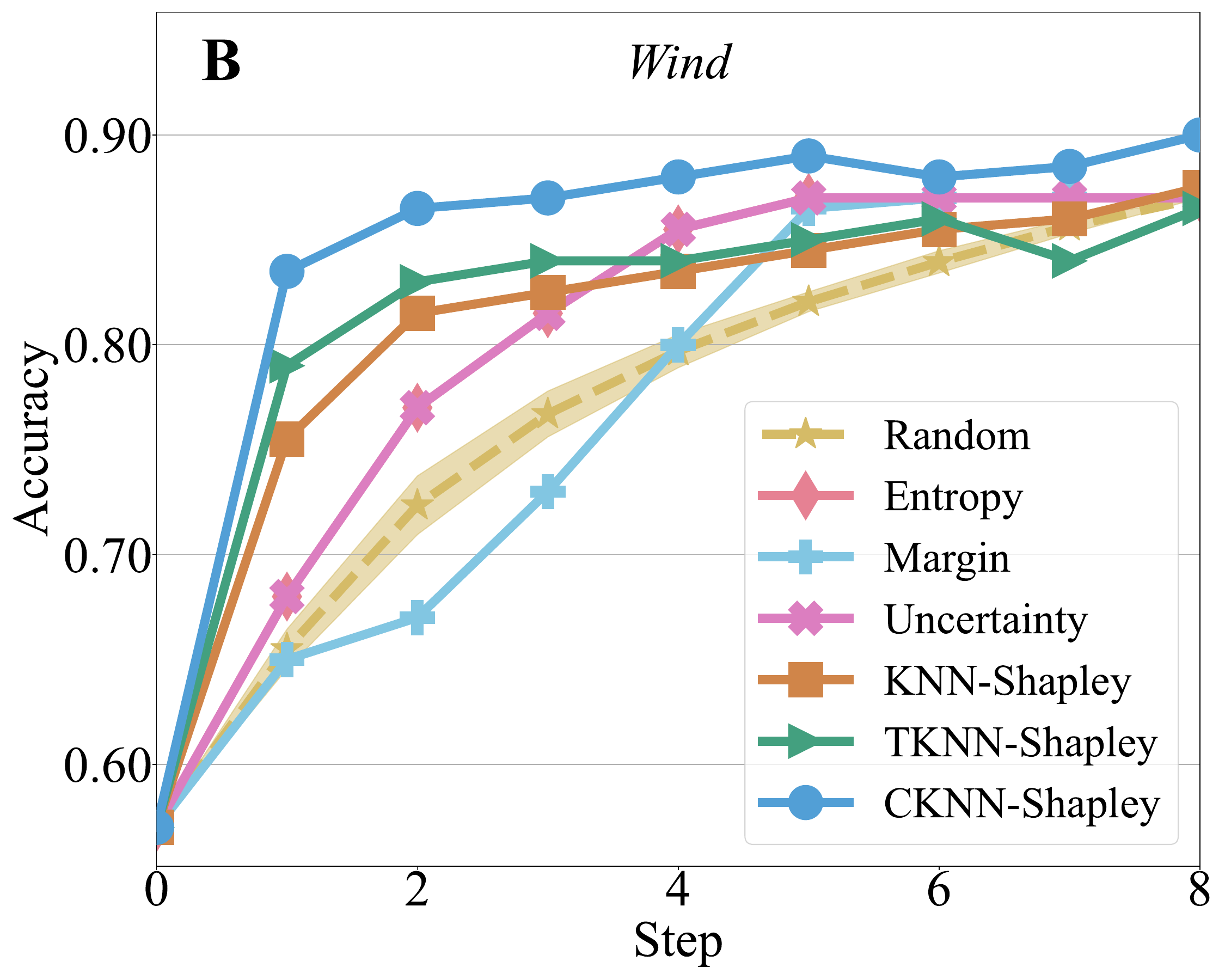} 
            }
        \subfigure{
            \includegraphics[width=0.31\textwidth]{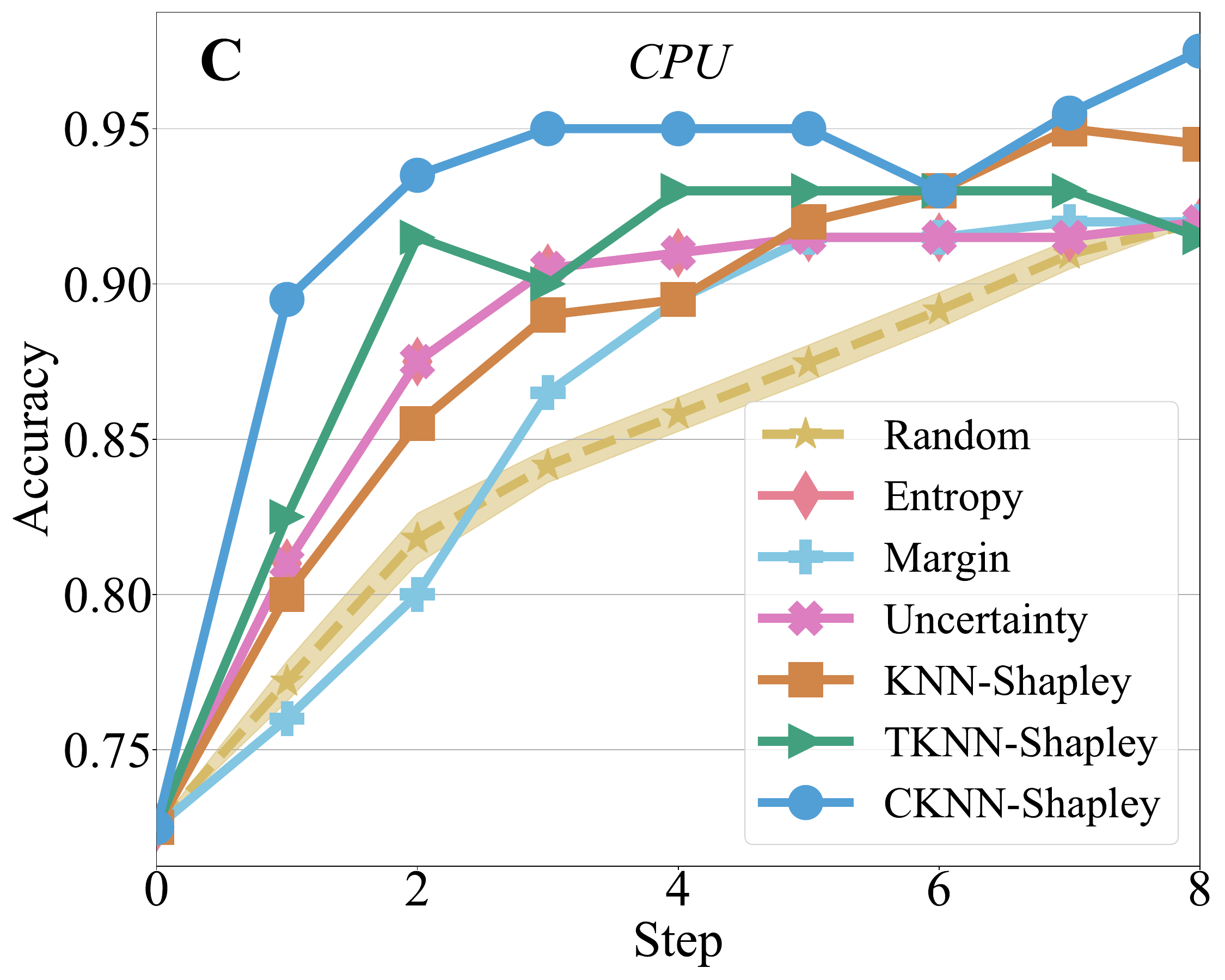} 
            } 
        % \subfigure{
        %     \includegraphics[scale=0.13]{figures/bar_plot_flip_ratio_legend.png} \vspace{-2mm}
        %     }
        \vspace{-4mm}
        \caption{Active learning of different annotation strategies on \textit{Pol}, \textit{Wind}, and \textit{CPU}, where the initial stage has 400 samples and each subsequent step annotates 200 samples into the training set. }
        \label{fig:indepth3}
        \vspace{-3mm}
    \end{figure*}

\textbf{Active Learning}.
To demonstrate the advantages of Shapley value in improved data utilization efficiency, we further evaluate our approach in the context of active learning. Specifically, to select the unlabeled data for labeling, we use a 2-layer fully connected neural network to fit the relationship between features and Shapley values of labeled training data. Then we use that neural network to predict Shapley values of unlabeled training data and choose the data with highest predicted Shapley values to label. In this experiment, we choose three chemical datasets —\textit{Pol}, \textit{Wind}, and \textit{CPU}. We segment these datasets into two parts, 20 percent of them are used as labeled data, other 80 percent are used as unlabeled data. The unlabeled data will be taken out 8 times through the prediction of neural network. Figure~\ref{fig:indepth3} shows the performance of three
KNN-Shapley-based data valuation approaches and three baselines of active learning (i.e., random sampling, entropy sampling~\cite{holub2008entropy}, margin sampling~\cite{balcan2007margin}, uncertainty sampling~\cite{nguyen2022measure}) on three chemical datasets. Random sampling involves the completely random selection of unlabeled samples for labeling; entropy sampling chooses samples with high prediction uncertainty, measured by entropy; margin sampling focuses on the gap between the two highest probabilities in the model's predictions; Uncertainty sampling prioritizes annotation of samples with high nearest neighbor inconsistency. From Figure~\ref{fig:indepth3}, CKNN-Shapley outperforms other methods and is more effective in identifying useful samples in the unlabeled data, extending the application context of Shapley value and providing a new way to select data for active learning.

\section{Conclusion}
In this paper, we revealed the value inflation of KNN-Shapley, which not only misidentifies a large portion of detrimental samples as beneficial, but also overestimates the value for the majority of samples. To address these issues, we proposed Calibrated KNN-Shapley to calibrate zero as the threshold for distinguishing detrimental samples from beneficial ones, via mitigating the negative effects of small training subsets when calculating data valuation. Through extensive experiments, we demonstrated the effectiveness of CKNN-Shapley in alleviating data valuation inflation and detecting detrimental samples. Furthermore, we extended our approach beyond conventional classification settings to the context of learning with mislabeled samples, online learning, and active learning.

%Bibliography
\bibliographystyle{unsrt}  
\bibliography{references}

\clearpage
\appendix

\section*{Appendix}
\section{Dataset}\label{app:dataset}
A comprehensive list of datasets and sources is summarized in Table \ref{tab:dataset_summary}. Following~\cite{wang2023threshold}, the validation data size we use is also 10\% of the training data size. For \textit{Pol}, \textit{Wind}, and \textit{CPU} datasets, we subsample the dataset to balance positive and negative labels. For the image dataset CIFAR10, we apply a ResNet50~\cite{he2016deep} that is pre-trained on the ImageNet dataset as the feature extractor. This feature extractor produces a 1024-dimensional
vector for each image. We also employ Sentence BERT~\cite{reimers2019sentence} as embedding models to extract features for the text classification dataset \textit{AGNews}, \textit{SST-2}, and \textit{News20}. This feature extractor also produces a 1024-dimensional
vector for each text instance. This method of using a foundation model as a feature extractor can make CKNN-SV well applied to data valuation of deep learning. For other datasets, we do not use any extractor.
\begin{table}[h]
    \centering
    \caption{A summary of datasets used in experiments}\vspace{-1mm}
    \resizebox{0.5\textwidth}{!}{
    \begin{tabular}{lccccc}
        \toprule
        \textbf{Dataset} & \textbf{\# Sample}  &\textbf{\#Dimension} &  \textbf{\#Class} & \textbf{Embeddings} \\
        \midrule
        \textit{MNIST} &  50000  & 784  & 10 &  None \\
        \textit{FMNIST} &  50000 & 784 &   10 & None \\
        \textit{CIFAR10}  & 50000 & 2048  & 10 & ResNet50 \\
        \textit{Pol} & 2000 & 48  & 2 & None \\
        \textit{Wind} & 2000 & 14& 2  & None \\
        \textit{CPU}  & 2000 & 21 & 2 &  None \\
        \textit{AGnews}  & 10000 & 384 & 4 & Sentence BERT \\
        \textit{SST-2}  & 10000& 384 & 2 & Sentence BERT \\
        \textit{News20}  & 10000 & 384 & 2 & Sentence BERT \\
        \bottomrule
    \end{tabular}}
    \label{tab:dataset_summary}
\end{table}
% A comprehensive list of datasets and sources is summarized in Table \ref{tab:dataset_summary}. Here we use 9 datasets for empirical evaluation. \textit{MNIST}, \textit{FMNIST}~\cite{xiao2017fashion}, and \textit{CIFAR10} are image datasets with 50,000 samples; \textit{Pol}, \textit{Wind}, and \textit{CPU}~\cite{wang2023threshold} are from the telecommunication, meteorology, and computer hardware domains, respectively, with 2,000 samples each; \textit{AGnews}, \textit{SST-2}~\cite{socher2013recursive}, and \textit{News20}~\cite{lang1995newsweeder} are text datasets with 10,000 samples. For the \textit{CIFAR10} and text datasets, we employ the ResNet50~\cite{he2016deep} and Sentence Bert~\cite{reimers2019sentence} as embedding for the KNN classifier, respectively. Features/embeddings of these nine datasets are from 14 to 2,048. The size of each dataset we use is shown in Table 7. For some of the datasets, we use a subset of the full set. The validation data size we use is 10\% of the training data size ~\cite{wang2023threshold}. 
% \subsubsection{Fixed Amount of Removed Samples}
% We conduct the experiments by removing 10, 20, 30 percent samples with smallest Shapley values, shown in the table \ref{tab:102030percent}. We can see our CKNN-Shapley achieves the best performance in the most cases under this setting. \\
\section{Additional Experiments}~\label{app:additional}
We provide additional experiments of our CKNN-Shapley in terms of exploration of misidentified samples, generalization on other classifiers, and various values of $T$.

\subsection{Misidentified Samples}
% Misidentified samples—both false positives (FP) and false negatives (FN)—on the overall model performance. As suggested, we report the number of false positives and false negatives in identifying detrimental samples across various data valuation methods in the table below. 

Our CKNN-Shapley method exhibits significantly lower false positives compared to other prevalent methods. Since the over-identification of samples as detrimental (FP) is a primary source of value inflation in data valuation, the number of FP samples is much larger than the number of FN samples in all three methods.

\begin{table*}[h]
\centering
\vspace{-4mm}
\caption{False positive and false positive of misidentified samples}\label{tab:false}\vspace{1mm}
\resizebox{0.98\textwidth}{!}{%
\begin{tabular}{p{3.5cm}|*{9}{>{\centering\arraybackslash}p{1.2cm}}|>{\centering\arraybackslash}p{1.2cm}}
\toprule
Method$\backslash$Datasets  & \textit{MNIST} & \textit{FMNIST} & \textit{CIFAR10} & \textit{Pol} & \textit{Wind} & \textit{CPU} & \textit{AGnews} & \textit{SST-2} & \textit{News20} & Sum. \\ 

 \midrule
 \multicolumn{8}{l}{False positive}\\
 \midrule
% LOO$^\#$ & - & \\  
KNN-Shapley& 47& 21 & 30 & 68 & 71 & 72 & 52 & 55 & 45 & 409\\ 
TKNN-Shapley& 17& 37 & 0 & 47 & 62 & 64 & 15 & 7 & 14 & 263 \\ 
CKNN-Shapley & 14& 9 & 7 & 21 & 28 & 22 & 22 & 8 & 17 & 148\\ 
 \midrule
 \multicolumn{8}{l}{False negative}\\
 \midrule
% LOO$^+$ & - & \\ 
KNN-Shapley& 0& 1 & 0 & 1 & 1 & 0 & 2 & 0 & 1 & 6\\ 
TKNN-Shapley& 41& 5 & 55 & 11 & 6 & 8 & 11 & 19 & 18 & 174\\ 
CKNN-Shapley& 0& 5 & 1 & 1 & 2 & 2 & 0 & 2 & 2 & 15\\ 
\bottomrule
\end{tabular}%
}
\end{table*}

\subsection{Generalization}
Traditional Shapley values have no constraint on the base classifier, but have to require retraining the base classifier $2^N$ times. Therefore, even for a small dataset, $2^N$ times model training is inflexible for the traditional Shapley values. KNN-Shapley and its variants including our CKNN-Shapley work a series of pragmatic tools computing Shapley values efficiently in a recursive manner with $O(N\log N)$ or $O(N)$ time complexity. For practice, if the base classifier is not KNN, a routine solution employs the KNN classifier as a surrogate. 

To test the generalization of our CKNN-Shapley on different models, we conduct extra experiments with non-KNN classifiers, including Multilayer Perceptron (MLP), Logistic Regression (LR), and Support Vector Machine (SVM). Specifically, we train the non-KNN classifiers with the whole training set, apply KNN-Shapley variants to identify detrimental samples, and finally retrain the non-KNN classifiers with the identified detrimental samples removed. Table \ref{tab:generalizability} reports the accuracy on three datasets. We can see that our CKNN-Shapley can effectively boost the performance of non-KNN classifiers in most cases, indicating a good generalization of CKNN-Shapley across different models.

\begin{table}[h]
\centering
\vspace{-2mm}
\caption{Generalizability on other classifier}\label{tab:classifier}\vspace{-1mm}
\resizebox{0.65\textwidth}{!}{%
\begin{tabular}{lccc}
\toprule
Method$\backslash$Datasets  & \textit{Pol} & \textit{Wind} & \textit{CPU} \\ 
 \midrule
Vanilla MLP & 0.9909 & 0.8735 & 0.9475\\
%MLP with negative Data-Shapley value  samples removed& 0.9919 & 0.8809 & 0.9535\\
MLP with negative KNN-Shapley value  samples removed & 0.9949 & 0.8939 & 0.9455\\ 
MLP with negative TKNN-Shapley value  samples removed & 0.8515 & 0.8269 & 0.9085\\ 
MLP with negative CKNN-Shapley value  samples removed & \textbf{0.9915} & \textbf{0.8979} & \textbf{0.9669}\\ 
\midrule
Vanilla LR & 0.8700 & 0.8500 & 0.9300\\
LR with negative KNN-Shapley value samples removed & 0.8800 & 0.8800& 0.9400\\ 
LR with negative TKNN-Shapley value samples removed & 0.8250 & 0.8250 & 0.9200\\ 
LR with negative CKNN-Shapley value samples removed & \textbf{0.8850} & \textbf{0.9350} & \textbf{0.9600}\\ 
  \midrule
  Vanilla SVM & \textbf{0.9650} & 0.8750 & 0.9400\\
SVM with negative KNN-Shapley value samples removed & 0.8650 & 0.8850& 0.9350\\ 
SVM with negative TKNN-Shapley value samples removed & 0.8450 & 0.8250 & 0.8900\\ 
SVM with negative CKNN-Shapley value samples removed & 0.9600 & \textbf{0.8850} & \textbf{0.9450}\\ 
\bottomrule
\end{tabular}%
}
\label{tab:generalizability}
\end{table}

\subsection{Rationality of $T$}
We consider two aspects when setting $T$. On one aspect, we believe the value inflation comes from the subset with too small sizes, which leads the sizes of subsets to be close to the whole training set (Large $T$). On another aspect, we expect the subsets to be diverse and the number of subsets to be large enough (Small $T$). By considering both, we give a default setting of $T = N - 2K$, which makes the subset size large enough and consider $2^{N-2K}\times (2^{2K}-1)$ different subsets. 

Moreover, we provide extra experiments of our CKNN-Shalpey with different values of $T$ below. In Table~\ref{tab: different t}, we set $T$ to be $0.95N, 0.75N, 0.5N$ and report the accuracy performance of removing samples with negative values (Larger value means better performance), threshold for distinguishing detrimental from beneficial samples (Closer to zero value means better performance), and misidentification ratio of detrimental samples (Smaller value means better performance). In general, the setting with $N-2K$ achieves the best average performance on all three metrics. For accuracy, the setting with $N-2K$ achieves the best performance on all datasets compared with other settings. For other metrics, the setting with $N-2K$ delivers competitive results. Therefore, we chose $T=N-2K$ as the default setting. 

\begin{table*}[h]
\centering
\vspace{-4mm}
\caption{Performance of CKNN-Shapley with different values of $T$}\label{tab:T}\vspace{1mm}
\resizebox{0.98\textwidth}{!}{%
\begin{tabular}{p{3.5cm}|*{9}{>{\centering\arraybackslash}p{1.2cm}}|>{\centering\arraybackslash}p{1.2cm}}
\toprule
Method$\backslash$Datasets  & \textit{MNIST} & \textit{FMNIST} & \textit{CIFAR10} & \textit{Pol} & \textit{Wind} & \textit{CPU} & \textit{AGnews} & \textit{SST-2} & \textit{News20} & Avg./Abs. \\ 
 \midrule
T & 0.9630 & 0.8444 & 0.5956 & 0.9400 & 0.8700 & 0.9200 & 0.9060 & 0.7270 & 0.6920 & 0.8287\\
 \midrule
 \multicolumn{8}{l}{Acc after remove samples with negative Shapley values}\\
 \midrule
% LOO$^\#$ & - & \\ 
N-2K & \textbf{0.9828} & \textbf{0.8884} & \textbf{0.7164} & \textbf{0.9700} & \textbf{0.9000} & \textbf{0.9700} & \textbf{0.9420} & \textbf{0.8980} & \textbf{0.7960}  & \textbf{0.8960}\\ 
0.95N  & 0.9698 & 0.8444 & 0.6556  & 0.9700 & 0.8850 & 0.9650 & 0.9270 & 0.8612 & 0.7700 & 0.8719\\ 
0.75N  & 0.9696 & 0.8444 & 0.6520 & 0.9650 & 0.8600 & 0.9450 & 0.9250 & 0.8463 & 0.7610 & 0.8632\\ 
0.5N  & 0.9630 & 0.8444 & 0.6498 & 0.9650 & 0.8650 & 0.9500 & 0.9250 & 0.8474 & 0.7590  & 0.8631\\ 
 \midrule
 \multicolumn{8}{l}{Threshold for distinguishing detrimental from beneficial samples by $\times$ $1e$$-4$}\\
 \midrule
% LOO$^+$ & - & \\ 
N-2K  & \textbf{0.0152} & \textbf{0.0109} & 0.1163 & -0.3059 & \textbf{-0.0245} & \textbf{-2.9412} & 1.1955 & \textbf{0.1604} & \textbf{0.6715} & \textbf{0.6046} \\ 
0.95N  & 0.3888 & 0.2097& 0.1389 & 0.3889& 3.6552 & 1.1062& \textbf{1.1398} & 1.7674 & 1.2325 & 1.1141\\ 
0.75N & 0.1970 & 0.2222& 0.1294 & \textbf{-0.1216}& 0.4348 & 3.8879 & 1.7923 & 0.6662 & 1.8334 & 1.0316\\ 
0.5N   & 0.2352 & 0.2177& \textbf{0.0718} & 0.4385& 4.3249 & 2.4663& 1.7139 & 0.8520 & 1.7043 & 1.3361 \\ 
 \midrule
 \multicolumn{8}{l}{Misidentification ratio of detrimental samples}\\
 \midrule
% LOO$^+$ & - & \\ 
N-2K & 0.2000 & \textbf{0.1686} & \textbf{0.0908} & \hspace{0.3em}\textbf{0.0000} & \hspace{0.3em}\textbf{0.0000} & \hspace{0.3em}\textbf{0.0000} & \textbf{0.1538} & \textbf{0.1143} & \textbf{0.1821} & \textbf{0.1011} \\ 
0.95N  & \textbf{0.1250} & 0.3333& 0.3736 & 0.1250& 0.0833 & 0.2727& 0.5172 & 0.2895 & 0.4737 & 0.2811\\ 
0.75N & 0.6325 & 0.4566& 0.3600 & 0.0000& 0.1000 & 0.4615& 0.5493 & 0.2500 & 0.5800 & 0.3767\\ 
0.5N  & 0.7107 & 0.4385& 0.3404 & 0.1111& 0.1429 & 0.3333& 0.6739 & 0.2432 & 0.5680 & 0.3985\\ 
\bottomrule
\end{tabular}%
}
\label{tab: different t}
\end{table*}

\section{Limitations and Broader Impact}\label{app:impact}
Our research focuses on overcoming challenges associated with value inflation in the application of KNN-Shapley for data valuation influence estimation. By introducing a method that enables accurate assessment of training samples' impact on model performance, we provide a tool for practitioners to determine the positive or negative effects of these samples. Through comprehensive testing across various problem scenarios, our Shapley value removal strategy has been proven superior to existing methods, enhancing model efficiency by eliminating harmful data points. Consequently, our contributions have the potential to drive substantial societal benefits, particularly as the use of more complex and expansive neural networks, like Large Language Models, becomes more prevalent. For the limitations, we do not provide a theoretical analysis on the selection of $T$, where we posit
that the choice of $T$ is contingent upon the dataset characteristic. To tackle this limitation, we recommend $T=N-2K$ as the default setting for its satisfactory performance on the datasets in our experiments. 

\section{Code and Reproducibility}\label{app:code}
% We provide our code, instructions, and implementation in an open-source repository: \href{https://github.com/yangziao56/Inflation_KNN-SV}{https://github.com/yangziao56/Inflation\_KNN-SV}. The experiments were conducted on a Linux (Ubuntu 20.04.6 LTS) server using NVIDIA GeForce RTX 4090 GPUs with 24GB VRAM running CUDA version 12.3 and driver version 545.23.08.
We will release the code soon.

\end{document}